\documentclass[10pt,twocolumn,letterpaper]{article}

\usepackage[pagenumbers]{iccv} % To force page numbers, e.g. for an arXiv version

\newcommand{\venueTT}[1]{{$_{\texttt{\text{#1}}}$}}

\definecolor{iccvblue}{rgb}{0.21,0.49,0.74}
\usepackage[pagebackref,breaklinks,colorlinks,allcolors=iccvblue]{hyperref}

\usepackage{hyperref}       % hyperlinks
\usepackage{url}            % simple URL typesetting
\usepackage{booktabs}       % professional-quality tables
\usepackage{amsfonts}       % blackboard math symbols
\usepackage{nicefrac}       % compact symbols for 1/2, etc.
\usepackage{microtype}      % microtypography

\usepackage{amsmath}
\usepackage{graphicx}
\usepackage{subcaption}
\usepackage{multirow}
\usepackage{array}
\usepackage{multirow}
\usepackage{ragged2e}
\usepackage{mathrsfs}
\usepackage{booktabs}
\usepackage{wrapfig}

\usepackage{bbm}

\def\paperID{2375}
\def\confName{ICCV}
\def\confYear{2025}

\title{GT-Loc: Unifying When and Where in Images Through a Joint Embedding Space}

\author{David G. Shatwell\textsuperscript{1} \quad Ishan Rajendrakumar Dave\textsuperscript{2} \quad Sirnam Swetha\textsuperscript{1} \quad Mubarak Shah\textsuperscript{1}\\
\textsuperscript{1}Center for Research in Computer Vision, University of Central Florida \quad \textsuperscript{2}Adobe\\
{\tt\small david.shatwell@ucf.edu \quad idave@adobe.com \quad swetha.sirnam@ucf.edu \quad shah@crcv.ucf.edu}
}

\begin{document}
\maketitle
\begin{abstract} 

Timestamp prediction aims to determine when an image was captured using only visual information, supporting applications such as metadata correction, retrieval, and digital forensics. In outdoor scenarios, hourly estimates rely on cues like brightness, hue, and shadow positioning, while seasonal changes and weather inform date estimation. However, these visual cues significantly depend on geographic context, closely linking timestamp prediction to geo-localization. To address this interdependence, we introduce GT-Loc, a novel retrieval-based method that jointly predicts the capture time (hour and month) and geo-location (GPS coordinates) of an image. Our approach employs separate encoders for images, time, and location, aligning their embeddings within a shared high-dimensional feature space. Recognizing the cyclical nature of time, instead of conventional contrastive learning with hard positives and negatives, we propose a temporal metric-learning objective providing soft targets by modeling pairwise time differences over a cyclical toroidal surface. We present new benchmarks demonstrating that our joint optimization surpasses previous time prediction methods, even those using the ground-truth geo-location as an input during inference. Additionally, our approach achieves competitive results on standard geo-localization tasks, and the unified embedding space facilitates compositional and text-based image retrieval.

\end{abstract}    
\section{Introduction}
\label{sec:intro}

\begin{figure}
  \centering
  \includegraphics[width=0.9\linewidth]{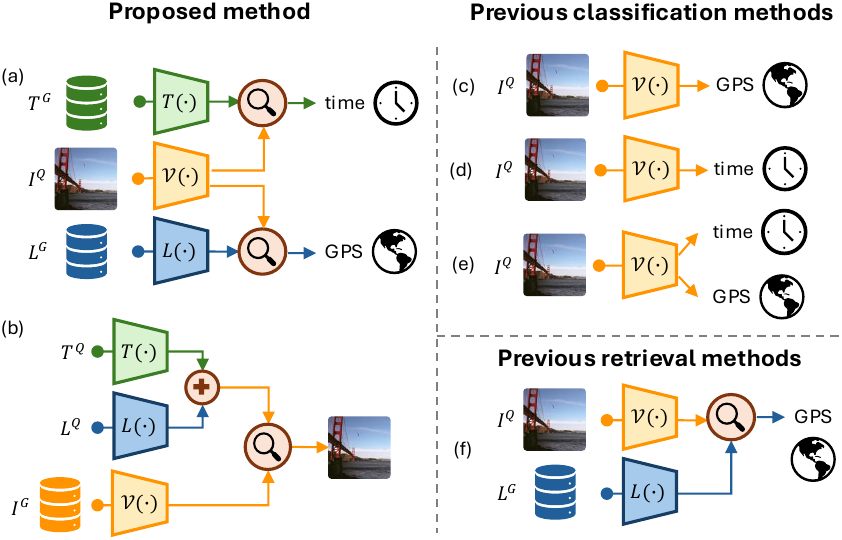}
  \caption{
    \textbf{GT-Loc: Our Unified Approach vs. Prior Methods}. By mapping image, location and time into a single multimodal embedding space, our method can be used for (a) simultaneous image-to-location and image-to-time retrieval, (b) composed geotemporal-to-image retrieval. In contrast, current methods are limited to only (c) location \cite{Haas_2024_CVPR, clark2023we}, (d) time \cite{salem2022timestamp} or (e) geo-temporal classification \cite{zhai2019learning} or (f) image-to-GPS retrieval \cite{klemmer2023satclip, vivanco2024geoclip}.
  }
  \label{fig:teaser}
\end{figure}

Estimating the capture time and geo-location of images is crucial for applications ranging from digital forensics to ecological studies and social media management. In digital forensics, accurate timestamps verify image authenticity and help detect manipulation, particularly when camera calibrations are suspect. This capability is essential for reconstructing events from timestamped images during accidents or natural disasters, providing critical information to first responders. Ecological studies benefit from time-ordered images to monitor changes in landscapes and wildlife, while precise timestamps in social media enhance content management and chronological sorting.

Despite its importance, predicting time from images presents several challenges because of the intricate relationship between temporal cues and location-specific factors. Time-of-day (ToD; \textit{i.e.}, hours) and time-of-year (ToY; \textit{i.e.}, months) manifest differently in images due to variables like scene brightness, shadows, weather, and seasonal changes, making it difficult to establish consistent patterns. The complexity of the task is further compounded as the visual appearance of specific hours varies substantially across different months and locations, influenced by the amount and relative exposure to sunlight. Additionally, the representation of months fluctuates across various latitudes, with regions near the equator experiencing relatively stable climate conditions year-round compared to regions at higher latitudes.

Most existing methods \citep{zhai2019learning, salem2022timestamp, padilha2022content} rely heavily on GPS metadata for accurate time estimation, while state-of-the-art geo-localization models (e.g., PIGEON \citep{Haas_2024_CVPR}, GeoCLIP \citep{vivanco2024geoclip}) excel at coarse location prediction. Yet predicting both time and location simultaneously without supplementary inputs remains unsolved, and few works tackle the complementary task of image retrieval (Figure~\ref{fig:teaser}). This challenge is further exacerbated by the absence of standardized datasets and evaluation protocols: many approaches use custom or poorly documented splits, hindering fair comparisons and consistent benchmarking across studies.

In this paper, we introduce \textit{GT-Loc}, a retrieval-based approach for joint time prediction and geo-localization. We conceptualize time prediction as a retrieval problem, representing time as a month-hour pair. A schematic diagram of our framework is shown in Figure~\ref{fig:framework}. Building upon the CLIP-initialized visual model, our goal is to learn a shared embedding space where we can align visual (image), time, and location modalities. For time prediction, we propose a novel time representation that considers the cyclical nature of months and hours over a toroidal manifold. We then project these representation into a multi-scale, high-dimensional time embedding using random Fourier features (RFFs) \citep{tancik2020fourier}. Next, to learn the alignment between the time and image embeddings, we explore several possibilities. Existing contrastive learning methods, including CLIP \citep{clip} and SimCLR \citep{chen2020simple}, use other batch instances as negative samples. Such a strategy succeeds in image-location losses used by Contrastive Spatial Pre-Training (CSP) \citep{pmlr-v202-mai23a}, GeoCLIP \citep{vivanco2024geoclip}, and SatCLIP \citep{klemmer2023satclip}, due to the significant variation of visual appearance with respect to geographical location. In contrast, image-time alignment suffers because temporal neighbors often look nearly identical: hours and months blend smoothly in appearance, so designating adjacent time points as negatives undermines effective alignment. Instead of defining positive-negative pairs as in contrastive learning, we propose a novel \textit{Temporal Metric Learning} approach, which encourages similarity between two instances based on their time difference. To build the target metric for our proposed loss, we use the toroidal distance between the times of each instance pair to consider the cyclic nature of time. This approach enhances performance without the need for explicit assignment of positive and negative samples, providing a more effective and efficient solution for time prediction.

By mapping the image, location, and time modalities into a unified feature space, our model is able to perform compositional retrieval tasks. For instance, given a specific time and location, it can efficiently retrieve all corresponding images from a gallery that closely match the specified criteria. 

In summary, our main contributions are the following:

\begin{itemize}

\item A framework for joint time-of-capture prediction and geo-localization by aligning the image, time and location embeddings in a shared multimodal feature space using contrastive learning.
\item The first retrieval-based method for time-of-capture prediction, where we propose a novel time representation as normalized month-hour pairs, considering its cyclic nature. 
\item A novel Temporal Metric Learning (TML) loss function for image-time alignment with soft targets, eliminating the need to assign positive and negative samples to the anchor. Since both hours and months are cyclic, we employ a toroidal distance instead of a regular $\ell_2$ distance which results in improved performance.
\item A new standard benchmarks for time prediction, demonstrating that our jointly optimized time-location method surpasses time-only optimized baselines and competes well with expert geo-localization methods. Our shared embedding space further facilitates downstream tasks like compositional and text-based retrieval.

\end{itemize}
\section{Related Work}
\label{sec:related}

\textbf{Time-of-capture prediction:} Time-of-capture prediction is a relatively new problem that has only been directly addressed by a handful of prior works. \citet{tsai2016photo} proposed a physically inspired method to infer the time of day by estimating the Sun's position and camera orientation, but their approach requires sky visibility and additional metadata, such as GPS coordinates and access to an external image database. In a different line of research, \citet{zhai2019learning} introduced a data-centric approach to learn geo-temporal image features. Their model uses an image, location, and time encoders to generate mid-level features, which are subsequently passed to a set of classifiers for predicting time and location as discrete classes. However, their evaluation shows that providing the location as an input is crucial for predicting the time of day with reasonable accuracy. Similarly, \citet{salem2022timestamp} proposed a hierarchical model to predict the month, hour, and week of capture, but this method also assumes known geo-location, limiting its real-world applicability. In contrast, our model relies solely on images to generate accurate time predictions using a retrieval approach in a continuous shared feature space, with resolution determined by a gallery of arbitrary size rather than discrete classes.

Other works have also explored the time-of-capture task indirectly. \citet{li2017you} presented an algorithm to verify image capture time and location by comparing the sun position, computed from the claimed time and location, with the actual sun position derived from shadow length and orientation. However, their approach assumes that latitude, time-of-day, and time-of-year are given, with only one potentially corrupted. \citet{padilha2022content} proposed a model for time-of-capture verification using a data-centric approach, involving four encoders for ground-level images, timestamps, geo-locations, and satellite images, fed into a binary classifier to predict time consistency. Similarly, \citet{salem2020learning} proposed a model to generate a global-scale dynamic map of visual appearance by matching visual attributes across images annotated with timestamps and GPS coordinates. Both \citet{padilha2022content} and \citet{salem2020learning} show qualitative results on time prediction, but are limited by their dependency on geo-location. In contrast, our method accurately estimates both GPS coordinates and timestamps using only images.

Several additional methods explore problems adjacent to time prediction. \citet{jacobs2007geolocating} proposed an algorithm for geo-locating static cameras by comparing temporal principal components of yearly image sequences with those from a gallery of known locations. For shadow detection, \citet{lalonde2012estimating} used a multi-stage method to extract pixel-wise ground-shadow features and find edges using CRF optimization, while \citet{wehrwein2015shadow} used illumination ratios to label shadow points in a 3D reconstruction and compute dense shadow labels in pixel space. Another series of works estimate the sun position for various downstream applications, such as computing camera parameters from time-lapses \citep{lalonde2010sun} and determining outdoor illumination conditions \citep{lalonde2012estimating, hold2017deep} from single images. Adapting these methods for time prediction would require additional metadata, which might not be available during inference. For example, even with correct sun position prediction, day of the year, geo-location, and compass orientation are needed to accurately predict the hour.

Although the above works contribute to time prediction, they either require additional input metadata like GPS coordinates, or are not reliable in the absence of specific temporal cues. In contrast, our proposed GT-Loc model aims to predict both ToY and ToD from a single image without relying on any additional metadata, making it more broadly applicable for real-time prediction tasks.

\textbf{Global geo-localization:} Geo-localization, the task of estimating the geographic coordinates of an image, has gained substantial popularity in recent years. Traditionally, geo-localization methods have adopted either a classification approach \citep{weyand2016planet, seo2018cplanet, vo2017revisiting, muller2018geolocation, pramanick2022world, kulkarni2024cityguessr} or an image retrieval approach \citep{regmi2019bridging, shi2019spatial, shi2020looking, toker2021coming, zhu2021vigor, zhu2022transgeo}. The classification approach divides the Earth into a fixed number of geo-cells or uses the city label, assigning the center coordinate of the selected class as the GPS prediction. However, this can result in significant errors depending on the size of the geo-cells, even when the correct class is selected. In contrast, the image retrieval approach compares a query image to a gallery and retrieves the image with the highest similarity. GeoCLIP \citep{vivanco2024geoclip} addresses the limitations of traditional approaches by framing global geo-localization as a GPS retrieval problem. It leverages the pretrained CLIP \citep{clip} ViT and employs contrastive learning to align image and location embeddings in a shared feature space. Other methods, such as PIGEON \citep{Haas_2024_CVPR}, use a hybrid strategy: first using image classification to identify the top-$k$ geo-cells with the highest probability, followed by a secondary retrieval stage for refinement within and across geo-cells. However, PIGEON's dependence on additional metadata for training—such as administrative boundaries, climate, and traffic—poses a significant limitation. Finally, recent methods such as Img2Loc \citep{zhou2024img2loc} have begun leveraging multimodal large language models (MLLMs) and retrieval-augmented generation (RAG) to achieve competitive geo-localization performance without the need for dedicated training. However, the effectiveness of these methods is highly dependent on the underlying MLLM and results in substantial inference overhead.

\textbf{Geo-spatial dual-encoder methods:} 
The success of CLIP has inspired to leverage its architecture for geo-spatial tasks. SatCLIP \citep{klemmer2023satclip} aligns satellite imagery and natural images in a shared feature space, enabling cross-modal retrieval and localization. In a similar fashion, \citet{zavras2024mind} proposed a method for aligning complementary remote sensing modalities beyond RGB with the CLIP encoders. Other works from \citet{pmlr-v202-mai23a} and \citet{Aodha_2019_ICCV}, employ a dual image-location encoder architecture to learn robust location representations from images. However, their ultimate goal is not to geo-locate images. Instead, they use the learned embeddings from the image encoder for downstream tasks, such as image classification. These methods demonstrate the effectiveness of dual-encoder architectures for geo-spatial problems, motivating our approach of using a triple encoder architecture for joint time-of-capture and location prediction.

\section{Method}
\label{sec:method}
\begin{figure*}
  \centering
  \includegraphics[width=0.9\linewidth]{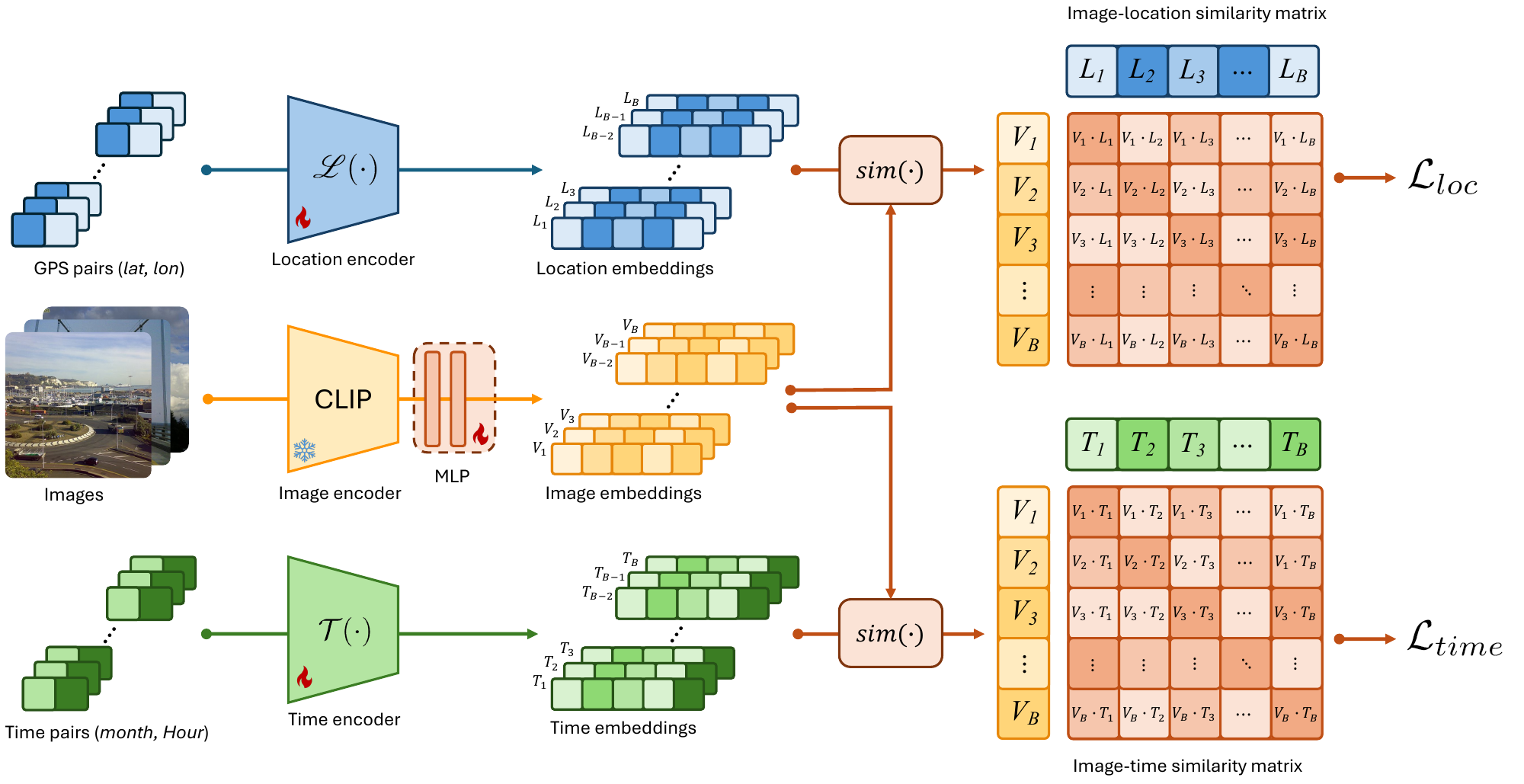}
  \caption{\textbf{Overview of GT-Loc}: GT-Loc uses an image encoder $\mathcal{V}(\cdot)$, location encoder $\mathscr{L}(\cdot)$ and time encoder $\mathcal{T}(\cdot)$ to generate a set of image $V_i$, location $L_i$ and time $T_i$ embeddings. 
  Leveraging the CLIP \cite{clip} pretrained ViT-L/14 as image encoder, we aim to align its image embedding to both location and time embeddings. 
  The image-location alignment is learned through a regular CLIP-like loss~\cite{vivanco2024geoclip} and the image-time alignment is learned through our proposed \textit{Temporal Metric Learning}.  
  }
  \label{fig:framework}
\end{figure*}

Given a training dataset $S_{train} = \{(I_i, G_i, D_i)\}_{i=1}^N$ consisting of image $I_i$, GPS coordinates $G_i$ and date-time $D_i$ triplets, our objective is to train a model that can simultaneously predict the location, time-of-day (ToD) and time-of-year (ToY) from unseen images.
Our GT-Loc method consists of three encoders: Image Encoder ($\mathcal{V}$), Location Encoder ($\mathscr{L}$), and Time Encoder ($\mathcal{T}$) as shown in Figure~\ref{fig:framework}. Both geo-localization and time prediction are framed as a retrieval problem. Given a query image $I^Q \in S_{eval}$, we compute an image embedding, $V^Q = \mathcal{V}(I^Q)$, using a pre-trained Vision Transformer. Similarly, given a gallery of latitude-longitude pairs, and a gallery of timestamps, we respectively compute galleries of location embeddings $L^Q_k = \mathscr{L}(G^Q_k)$ and time embeddings $T^Q_k = \mathcal{T}(D^Q_k)$. In order to predict the location and time, the image embedding is compared against both galleries. The GPS and timestamp with the highest cosine similarity to the image are selected as predictions. 

A fundamental prerequisite for retrieval is the alignment of image, location, and time modalities within a shared multimodal embedding space. To achieve this, our framework is optimized using two multimodal alignment objectives: (1) Image-Location alignment, and (2) Image-Time alignment. Furthermore, to capitalize on large-scale visual pretraining, we employ the pretrained CLIP ViT-L/14 as our image encoder, projecting it into our shared embedding space using a trainable multi-layer perceptron (MLP). 

\subsection{Image-Location Alignment}
We adopt GeoCLIP for the image-location modality alignment. Given a latitude-longitude pair $G_i$, it first uses Equal Earth Projection (EEP) to mitigate the distortion of the standard GPS coordinate system and provide a more accurate representation $G_i'$. Then, Random Fourier Features (RFF) are used to map the 2D representation into a rich high-dimensional representation at three scales ($M$) using projection matrices $\gamma(\cdot)$ with different frequencies $\sigma_i \in \left\{ 2^0,2^4,2^8 \right\}$. Lastly, the RFFs are passed to a set of MLPs $f_i$ and added together, forming a single multi-scale feature vector. This can be mathematically expressed as the following equation:
\begin{equation}
L_i = \mathscr{L}(G_i) = \sum_{i=1}^M f_i(\gamma(EEP(G_i), \sigma_i)).
\label{eq:gps_enc}
\end{equation}

Next, to compute the image-location contrastive loss $\mathcal{L}_{loc}^{i}$, we consider a set of $P$ augmented image $V_{ij}$ and location $L_{ij}$ embeddings ($j \in {1, \dots, P}$). For the batch with size $B$ with $S$ additional location embeddings stored in a continually updated dynamic queue, and temperature $\tau$, the loss is given by:
\begin{equation}
    \mathcal{L}_{loc}^{(i)} = -\sum_{j=1}^P \log \left( \frac{\exp(V_{ij} \cdot L_{ij} / \tau)}{\sum_{k=1}^{B+S} \exp(V_{kj} \cdot L_{kj} / \tau)} \right).
\label{eq:gps_loss}
\end{equation}

\begin{figure}
  \centering
  \includegraphics[width=\linewidth]{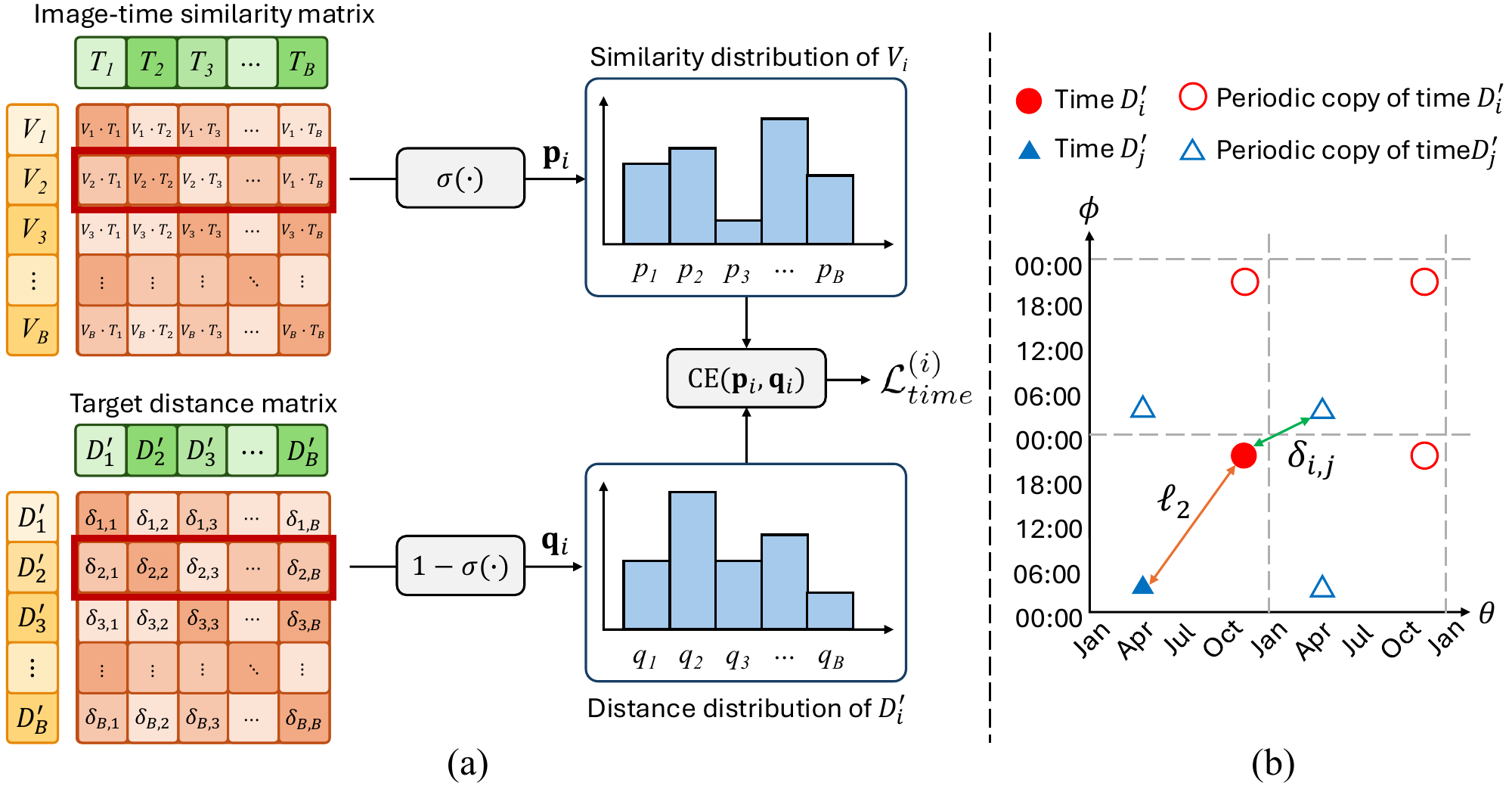}
  \caption{(a) \textbf{Proposed Temporal Metric Learning loss $\mathcal{L}_{time}$:} 
  We compute the \textbf{image-time similarity matrix} by taking the cosine distance between the image and time embedding of the all instances of the batch. We then obtain the \textbf{target distance matrix} by computing the cyclic toroidal time difference between each pair.   
  As shown in the red highlighted box, we take the $i$th row of both matrices and normalize them using the softmax function $\sigma(\cdot)$ and $1-\sigma(\cdot)$ respectively, resulting in two probability mass functions $\mathbf{p}_i$ and $\mathbf{q}_i$. The loss is then given by the cross-entropy (CE) between $\mathbf{p}_i$ and $\mathbf{q}_i$.       
   \textbf{(b) Our proposed distance metric} $\delta_{i,j}$: Assume we have two normalized month-hour pairs $D_i'=(\theta_i,\phi_i)$ and $D_j'=(\theta_j,\phi_j)$. Since month and hours are periodic, they repeat infinitely in both directions of the $\theta$-$\phi$ plane. Using the $\ell_2$ distance overestimates the real distance between the two times, but our proposed toroidal time distance $\delta_{i,j}$ is able to provide a correct estimate by considering the minimum distance between $D_i'$ and the periodic copies of $D_j'$.}
  \label{fig:metric_loss}
\end{figure}

\subsection{Time Representation}
The capture time of an image is usually a Unix timestamp, an integer tracking the seconds (or milliseconds) elapsed since January 1, 1970.
We discard the year information from the timestamp, as predicting the year is beyond the scope of this work, and instead focus on Time-of-Year (ToY; \textit{i.e.}, month) and Time-of-Day prediction (ToD; \textit{i.e.}, hour). Both ToY and ToD are cyclical, with periods of 12 months and 24 hours, respectively. To convert the Unix timestamp $U_i$ into a time representation that focuses on the months and hours, we transform it into a date-time tuple $D_i = \mathrm{\texttt{unix2tuple}}(U_i) = (m_i, d_i, H_i, M_i, S_i)$ with the month, day, hour, minute, and second. From this tuple, we then compute a new time representation composed of the  normalized cyclic month-hour pair $D_i'=(\theta_i, \phi_i)$ using
\begin{equation}
    \begin{aligned}
\theta_i = \frac{1}{12} \left( (m_i - 1) + \frac{(d_i - 1)}{\mathcal{D}(m_i)} \right),
    \end{aligned}
\label{eq:time_rep_theta}
\end{equation}
\begin{equation}
    \begin{aligned}
\phi_i = \frac{1}{24} \left( H_i + \frac{M_i}{60} + \frac{S_i}{3600} \right),
    \end{aligned}
\label{eq:time_rep_phi}
\end{equation}

\noindent where $\mathcal{D}(m_i)$ is the number of days in month $m_i$. We represent the sequence of operations to convert a Unix timestamp to normalized cyclic month-hour pair as $D_i' = \mathrm{\texttt{unix2cyclic}}(U_i)$.

\textbf{Time encoding:}
By representing time as a pair of real-valued numbers, the problem of time prediction becomes similar in nature to geo-localization. Instead of retrieving the latitude-longitude pair with the highest similarity, we are interested in retrieving a month-hour pair. 
Thus, our time encoder ($\mathcal{T}$) follows the exact same architecture as the location encoder ($\mathscr{L}$). Similar to Eq.~\ref{eq:gps_enc}, the time embedding is obtained from the proposed time representation using the following equation:
 
\begin{equation}
T_i = \mathcal{T}(U_i) = \sum_{i=1}^M f_i(\gamma(\mathrm{\texttt{unix2cyclic}}(U_i), \sigma_i)).
\label{eq:time_enc}
\end{equation}

\subsection{Temporal Metric Learning}

Visual features associated with location typically exhibit significant variation due to cultural, socio-economic and environmental factors, leading to abrupt changes in visual embeddings with respect to spatial distances. For instance, two neighborhoods within the same city often appear notably different. Therefore, standard contrastive objectives that clearly distinguish positives and negatives are suitable for aligning location and image embeddings (Eq.~\ref{eq:gps_loss}). In contrast, visual cues change more smoothly and continuously over time. As a result, labeling temporally adjacent samples as negatives, as is common in standard contrastive losses, can hinder effective learning. This makes it challenging to define explicit positives and negatives for time-image alignment. To address this, we propose a metric-learning objective called \textit{Temporal Metric Learning} (TML), explicitly designed to leverage the smooth and cyclic nature of time by aligning instance similarity proportionally to the temporal difference.

Let's consider the image embeddings $\left\{ V_i \right\}_{i=1}^{B}$ and time embeddings $\left\{ T_i \right\}_{i=1}^{B}$. Instead of defining a set of positive and negative pairs for each embedding, we assign soft targets inversely proportional to the difference between the time associated to the image and time embeddings. Since months and hours are cyclical, using the regular $\ell_2$ distance between two normalized month-hour pairs $(\theta_i, \phi_i)$ and $(\theta_j, \phi_j)$ in an Euclidean space results in overestimated distance values, as shown in Figure~\ref{fig:metric_loss}(b). This problem can be solved by mapping the normalized month-hour pairs into the surface of a toroidal manifold, resulting in the new distance $\delta_{i,j}$:
\begin{equation}
    \label{eq:t_dist}
\delta_{i,j} = \sqrt{\sum_{\alpha \in \left\{\theta,\phi\right\}}\min(1 - |\Delta \alpha_{i,j}|, |\Delta \alpha_{i,j}|)^2}.
\end{equation}

\noindent where $\Delta \alpha_{i,j}=\alpha_{i}-\alpha_{j}$. Then, for each anchor image embedding $V_i$, we compute a vector $\mathbf{p}_i$ with the normalized cosine similarity scores with respect to all time embeddings $T_j$ in the batch. Similarly, we compute the vector $\mathbf{q}_i$ with the normalized time difference between the anchor time and other times in the batch as follows:
\begin{equation}
    \mathbf{p}_{i}[j] = \frac{\exp \left( V_i \cdot T_j / \tau \right)}{\sum_{k=1}^B \exp \left( V_i \cdot T_k / \tau \right)}, \quad j \in [B],
\end{equation}

\begin{equation}
    \mathbf{q}_{i}[j] = 1 - \frac{\exp \left( \delta_{i,j} \right)}{\sum_{k=1}^B \exp \left( \delta_{i,k} \right)}, \quad j \in [B].
\end{equation}

Since $\mathbf{p}_i$ and $\mathbf{q}_i$ are normalized, they have the same characteristics as probability mass functions. Thus, we define the image-time contrastive loss by computing the cross-entropy ($\mathrm{CE}$) between $\mathbf{p}_i$ and $\mathbf{q}_i$:
\begin{equation}
    \mathcal{L}_{time}^{(i)} = \mathrm{CE}(\mathbf{p}_i,\mathbf{q}_i).
\end{equation}

The final training objective is defined as the sum of the image-location and image-time objectives, defined in Equations \ref{eq:gps_loss} and \ref{eq:time_loss}:
\begin{equation}
    \mathcal{L}^{(i)} = \mathcal{L}_{loc}^{(i)} + \mathcal{L}_{time}^{(i)}.
    \label{eq:time_loss}
\end{equation}

\begin{table}[t]
\centering
\setlength{\aboverulesep}{0pt}
\setlength{\belowrulesep}{0pt}
\caption{\textbf{Zero-shot time prediction} on the \textit{unseen} cameras of SkyFinder dataset. Rows marked by * indicate methods we replicate, closely adhering to the protocols outlined by prior work.}
\label{tab:time-pred}
\begingroup
\resizebox{\linewidth}{!}{
\setlength{\tabcolsep}{3pt}
\begin{tabular}{lccc} 
\toprule
\textbf{Method}  & \textbf{Month} & \textbf{Hour} & \textbf{TPS} $\uparrow$ \\
\textbf{} & \textbf{Error} $\downarrow$ & \textbf{Error} $\downarrow$ & \textbf{} \\

\midrule

\citet{zhai2019learning}* & 2.46 & 3.18 & 65.48 \\
\citet{padilha2022content}* & 1.62 & 3.61 & 71.42 \\
\citet{salem2022timestamp}* & 2.72 & 3.05 & 63.25 \\

\citet{zhai2019learning}* w/ CLIP & 1.65 & 3.14 & 73.20 \\
\citet{padilha2022content}* w/ CLIP & 1.62 & 2.98 & 74.06 \\
\citet{salem2022timestamp}* w/ CLIP & 1.56 & 2.87 & 75.02 \\

\midrule

CLIP + cls. head & 1.61 & 3.17 & 73.37 \\
CLIP + reg. head & 1.55 & 3.06 & 74.33 \\
DINOv2 + cls. head & 2.24 & 3.74 & 65.61 \\
DINOv2 + reg. head & 2.12 & 3.53 & 67.49 \\
OpenCLIP + cls. head & 1.66 & 3.43 & 71.87 \\
OpenCLIP + reg. head & 1.72 & 3.34 & 71.75 \\

\midrule

TimeLoc & \underline{1.52} & \underline{2.84} & \underline{75.49} \\

\rowcolor{yellow!30!orange!30}
  \textbf{GT-Loc (Ours)} & \textbf{1.40} & \textbf{2.72} & \textbf{77.00} \\

\bottomrule
\end{tabular}
}
\endgroup
\end{table}

\subsection{Inference}  

After training, the image, location, and time modalities share a unified embedding space. To predict the capture time and location of a query image $I^Q$, we compute the cosine similarity between its image embedding $V^Q$ and the embeddings within the time gallery $T^G$ and the location gallery $L^G$. The predicted time and location correspond to the gallery elements with the highest cosine similarities. For a visual representation of this inference process, refer to Supplementary Section~\ref{sec:m_infer}.
\section{Experiments}
\label{sec:exp}

We evaluate GT-Loc on both time-of-capture and geo-localization tasks, conduct ablation studies to justify key design choices, and assess robustness by measuring performance under limited training data and noisy annotations.

\textbf{Datasets and evaluation details}:
For training, we use two existing datasets: MediaEval Placing Tasks 2016 (MP-16) \citep{mp16} and Cross-View Time (CVT) \citep{salem2020learning}. MP-16 consists of 4.72 Million images from Flickr annotated only with GPS coordinates. CVT originally consists of 206k geo-tagged smartphone pictures from the Yahoo Flickr Creative Commons 100 Million Dataset \citep{thomee2016yfcc100m} and 98k images from static outdoor webcams of the SkyFinder Dataset \citep{mihail2016sky}. Our zero-shot evaluation benchmark is constructed using a subset of images from SkyFinder that are not seen during training.

Geo-localization performance is evaluated by measuring the geodesic distance between the real and predicted GPS coordinates, and then computing the ratio of images that are correctly predicted within a threshold. Time prediction performance is evaluated by measuring the mean absolute ToY ($E_{ToY}$) and ToD ($E_{ToD}$) errors between the ground-truth and predicted times. We also report an overall \textbf{Time Prediction Score (TPS)} that combines both errors into a single metric. We compute the TPS based on our proposed cyclical time difference using the following equation:
\begin{equation}
    \label{eq:tps}
TPS = 1 - \sqrt{\frac{\widetilde{E}_{ToY}^2 + \widetilde{E}_{ToD}^2}{2}},
\end{equation}
where $\widetilde{E}_{ToY}, \widetilde{E}_{ToD} \in [0,1]$ are the normalized time errors. A TPS of 1 represents a perfect prediction, while 0 represents the maximum possible cyclic error of 12 hours and 6 months. Please, refer to Supplementary Sections~\ref{sec:impl_supp} to \ref{sec:data_details} for more details about the dataset, architecture and training protocol.
\begin{table}[t]
\centering
\caption{\textbf{Geo-localization accuracy} on Im2GPS3k \& GWS15k datasets, reported on the ratio of samples that are correctly predicted under a distance threshold of 1 km radius.}
\label{table:geoloc_trimmed}
\begingroup
\resizebox{\linewidth}{!}{
\setlength{\tabcolsep}{3pt}
\begin{tabular}{lcc} 
\toprule
\textbf{Method}         & \textbf{Im2GPS3k} & \textbf{GWS15k}  \\ 
\hline
{[}L] kNN, sigma=4~\venueTT{ICCV'17\citep{vo2017revisiting}} & 7.2               & -                \\
PlaNet~\venueTT{ECCV'16\citep{weyand2016planet}}             & 8.5               & -                \\
CPlaNet~\venueTT{ECCV'18\citep{seo2018cplanet}}            & 10.2              & -                \\
ISNs~\venueTT{ECCV'18\citep{muller2018geolocation}}              & 10.5              & 0.1              \\
Translocator~\venueTT{ECCV'22\citep{pramanick2022world}}       & 11.8              & 0.5              \\
GeoDecoder~\venueTT{CVPR'23\citep{clark2023we}}          & 12.8              & 0.7              \\
GeoCLIP~\venueTT{NeuRIPS'24\citep{vivanco2024geoclip}}            & 14.11              & 0.6              \\
PIGEOTTO~\venueTT{CVPR'24\citep{Haas_2024_CVPR}}         & 11.3              & 0.7              \\
\textcolor{gray}{Img2Loc(LLaVA)~\venueTT{SIGIR'24\citep{zhou2024img2loc}}}    & \textcolor{gray}{8.0}                 & \textcolor{gray}{-}                \\ 
\hline
\rowcolor{yellow!30!orange!30}
\textbf{GT-Loc (Ours)}                  & \textbf{14.41}     & \textbf{0.88}     \\
\bottomrule
\end{tabular}
}
\endgroup
\end{table}

\subsection{Comparison with Prior Methods}

\noindent\textbf{Time-of-capture prediction:}
We present our time prediction results in Table \ref{tab:time-pred}. Given the significant reproducibility challenges identified in prior work (see Supplementary Section \ref{sec:supp_limitations}), we selected three representative baseline methods for a fair comparison with GT-Loc. The first baseline is the triple encoder architecture from \citet{zhai2019learning}, which closely resembles our approach. Although they did not release code, pretrained weights, or exact dataset splits, their methodology is well-documented, allowing us to closely replicate their protocol. The second baseline is \citet{padilha2022content}, chosen due to its recent publication, public source code availability, and use of the CVT dataset—common to our evaluation. However, they do not provide the specific cross-camera split of CVT, and their original work only offers qualitative results. We extend their evaluation quantitatively for direct comparison. The final baseline is \citet{salem2022timestamp}, selected for its use of the SkyFinder dataset, with clear experimental procedures enabling straightforward replication despite the absence of source code. Importantly, all these baseline methods rely on geo-location metadata as input, unlike GT-Loc, which exclusively uses visual cues. Additionally, each baseline fully trains their models using different backbone architectures: \citet{zhai2019learning} employs InceptionV2, while both \citet{padilha2022content} and \citet{salem2022timestamp} use DenseNet-121. To ensure fairness, we also evaluate scenarios in which these models use a frozen CLIP ViT-L/14 backbone, aligning closely with GT-Loc.

Our second set of baselines explores various frozen backbone architectures (e.g., CLIP ViT-L/14, DINOv2 ViT-L/14, and OpenCLIP ViT-G/14), paired with an MLP of matching capacity to our image encoder, and either regression or classification heads. These baselines illustrate the effectiveness of our retrieval-based approach compared to standard classification and regression techniques.

\begin{figure}[h]
  \centering
\includegraphics[width=\linewidth]{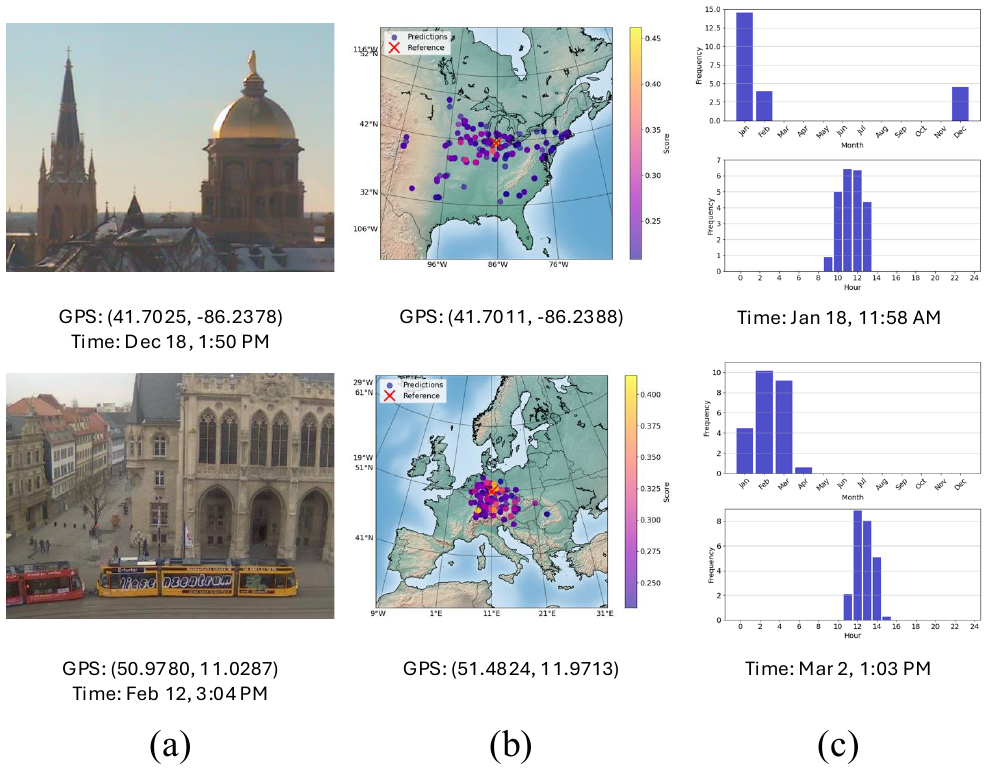}
  \caption{(a) Sample images for two cameras of the SkyFinder test set with the ground truth location and capture time. (b) Spatial distribution of the predicted GPS coordinates colored by the cosine similarity between the location and image embeddings. (c) Histogram of the top-1k retrieved months and hours, weighted by the cosine similarity between the image and time embeddings (supp. Eq. \ref{eq:supp_time_hist}). Both the top-1 predicted location and time are shown below the distributions.}
  \label{fig:qual}
\end{figure}

Lastly, we introduce TimeLoc, a robust baseline utilizing only the image and time encoders from GT-Loc, analogous to GeoCLIP \cite{vivanco2024geoclip} and SatCLIP \cite{klemmer2023satclip} but specifically tailored for time prediction. Experimental results clearly demonstrate that GT-Loc achieves superior accuracy for both Time-of-Year (ToY) and Time-of-Day (ToD) predictions without additional metadata. Moreover, jointly training for time prediction and geo-localization enriches the learned temporal representations. In Figure \ref{fig:qual} , we show two qualitative examples from unseen cameras of the SkyFinder dataset, highlighting GT-Loc's effectiveness in simultaneously predicting time and geo-location.

\begin{table}[ht]
\centering
\setlength{\aboverulesep}{0pt}
\setlength{\belowrulesep}{0pt}
\caption{\textbf{Ablations} for time prediction performance with different loss functions. $\ell_2$ refers to the Euclidean distance, while cyclic refers to the distance over the toroidal manifold.}
\label{tab:losses}
\begin{tabular}{lccc} 
\toprule
\textbf{Loss}  & \textbf{Month} & \textbf{Hour} & \textbf{TPS} $\downarrow$ \\
\textbf{Function} & \textbf{Error} $\downarrow$ & \textbf{Error} $\uparrow$ & \textbf{} \\
\midrule
CLIP \cite{clip}           & 1.71            & 3.51           & 71.12 \\
RnC \cite{rnc}       & 1.87            & 2.96           & 71.89 \\
SimCLR \cite{chen2020simple} & \underline{1.50}             & 3.40            & 73.28 \\
\midrule
TML ($\ell_2$)                    & 1.53            & \underline{2.74}           & \underline{75.88} \\
\rowcolor{yellow!30!orange!30}
\textbf{TML (Cyclic)}                & \textbf{1.40}             & \textbf{2.72}           & \textbf{77.00} \\
\bottomrule
\end{tabular}
\end{table}

\vspace{-1em}

\noindent\textbf{Geo-localization:} Table \ref{table:geoloc_trimmed} presents the geo-localization performance of our model compared to recent expert methods. Overall, GT-Loc delivers competitive results compared against models like GeoCLIP on both datasets. It also outperforms the LLaVA-based Img2Loc variant, despite using at least 15 times fewer parameters.

\subsection{Ablation Studies}

In this section, we present additional experiments to evaluate the effectiveness of our proposed Temporal Metric Loss. Furthermore, in the supplementary material (Section \ref{sec:ablations_supp}), we explore alternative image backbones (\textit{i.e.}, DINOv2 \citep{oquab2023dinov2}, OpenCLIP \citep{ilharco_gabriel_2021_5143773}), different time encoders (\textit{i.e.}, Time2Vec \citep{kazemi2019time2vec}, Circular Decomposition \citep{Aodha_2019_ICCV}), and various temporal resolutions to provide additional insights and justification for our design choices.

We evaluate several alternative loss functions for time prediction. First, we implement a basic CLIP-based loss \citep{clip}. Next, we explore a geo-localization-style contrastive loss that uses a dynamic queue and applies a false negative mask, excluding samples close to the anchor based on a specified threshold. We also experiment with the Rank-N-Contrast loss introduced by \citet{rnc}, specifically designed for regression tasks by ranking samples according to their distances. Finally, we compare our proposed loss function against a variant using standard $\ell_2$ distance rather than cyclic temporal distance. The results in Table \ref{tab:losses} demonstrate that our Temporal Metric Loss significantly outperforms all other evaluated losses for both Time-of-Day (ToD) and Time-of-Year (ToY) predictions.

\begin{table}[ht]
\centering
\setlength{\aboverulesep}{0pt}
\setlength{\belowrulesep}{0pt}
\caption{\textbf{Impact of Limited Data} on the Robustness of Time Prediction}
\label{tab:robustness-limited-data}
\begin{tabular}{lccc} 
\toprule
\textbf{Data} & \textbf{Month} & \textbf{Hour} & \textbf{TPS} $\uparrow$ \\
\textbf{Availability} & \textbf{Error} $\downarrow$ & \textbf{Error} $\downarrow$ & \textbf{} \\
\midrule
100\%                  & 1.40           & 2.72           & 77.00 \\
50\%                   & 1.69           & 2.83           & 74.02 \\
10\%                   & 1.70           & 2.94           & 73.51 \\
5\%                    & 1.89           & 2.86           & 72.07 \\
\bottomrule
\end{tabular}
\end{table}

\vspace{-1em}

\begin{table}[ht]
\centering
\setlength{\aboverulesep}{0pt}
\setlength{\belowrulesep}{0pt}
\caption{\textbf{Ablations} for robustness to label noise for time-prediction.}
\label{tab:robustness-noise}
\begin{tabular}{lccc} 
\toprule
\textbf{Label} & \textbf{Month} & \textbf{Hour} & \textbf{TPS} $\uparrow$ \\
\textbf{Noise ($\sigma$)} & \textbf{Error} $\downarrow$ & \textbf{Error} $\downarrow$ & \textbf{} \\
\midrule
0                  & 1.40           & 2.72           & 77.00 \\
1                  & 1.52           & 2.71           & 76.00 \\
2                  & 1.75           & 2.74           & 73.81 \\
3                  & 2.16           & 2.72           & 69.92 \\
\bottomrule
\end{tabular}
\end{table}

\subsection{Robustness of GT-Loc}

Timestamp‐prediction datasets are often scarce or plagued by missing and noisy metadata, so robustness is essential. We first measure performance as we shrink the training set from 100\% to 5\% in four stages, then simulate label noise by adding Gaussian perturbations with standard deviation between $\sigma \in [0,3]$ months and hours to the training annotations. As Tables \ref{tab:robustness-limited-data} and \ref{tab:robustness-noise} show, GT-Loc’s errors rise only modestly by 0.3 months and 0.22 hours, even with just 5\% of the data, and it remains stable up to $\sigma=2$, with significant degradation appearing only at $\sigma\geq3$. These results confirm GT-Loc’s resilience under realistic data constraints.

\subsection{Compositional Image Retrieval}

In this section, we evaluate GT-Loc’s capability for compositional image retrieval tasks, leveraging its unified multimodal embedding space. Specifically, we explore retrieving images based on joint queries consisting of both location and time information. Given query location $L^Q$ and time $T^Q$, we generate a multimodal representation by averaging their embeddings, inspired by multimodal retrieval methods proposed in prior works \citep{shvetsova2022everything, swetha2023preserving}.

To provide meaningful baselines for comparison, we selected the method by \citet{zhai2019learning}, which is conceptually similar but was originally intended only for time and location classification tasks. Since their original work did not consider compositional retrieval explicitly, we adapt their model to our retrieval scenario as described in Supplementary section \ref{sec:compositional}. We evaluate retrieval performance using recall metrics at ranks 1, 5, and 10. A retrieved image is considered correct if its ground truth timestamp is within one hour and one month, respectively, of the query, and if its location is within 25 km.

Table \ref{tab:time-pred} shows quantitative results for these baselines. GT-Loc consistently achieves higher recall across all ranks, significantly outperforming other methods. This indicates that our unified retrieval-based embedding approach effectively encodes joint geo-temporal information, providing richer representations for retrieval tasks. Qualitative examples demonstrating GT-Loc’s compositional retrieval capabilities are shown in Supplementary section \ref{sec:compositional}.

\begin{table}[ht]
\centering
\setlength{\aboverulesep}{0pt}
\setlength{\belowrulesep}{0pt}
\caption{\textbf{Zero-shot composed retrieval ($T+L \rightarrow I$)} on the unseen cameras of the SkyFinder dataset.}
\label{tab:composed-ret}
\begingroup
\resizebox{\linewidth}{!}{
\begin{tabular}{lccc} 
\toprule
\textbf{Method} & \textbf{R@1} & \textbf{R@5} & \textbf{R@10} \\
\midrule
\citet{zhai2019learning}* & 0.91 & 7.81 & 13.61 \\
\citet{zhai2019learning}* w/ CLIP & 2.58 & 16.22 & 29.46 \\
\midrule
\rowcolor{yellow!30!orange!30}
\textbf{GT-Loc} & \textbf{6.69} & \textbf{24.58} & \textbf{38.54}   \\
\bottomrule
\end{tabular}
}
\endgroup
\end{table}

\vspace{-1em}
\subsection{Qualitative results using text queries}
\begin{figure}[h]
  \centering
  \includegraphics[width=\linewidth]{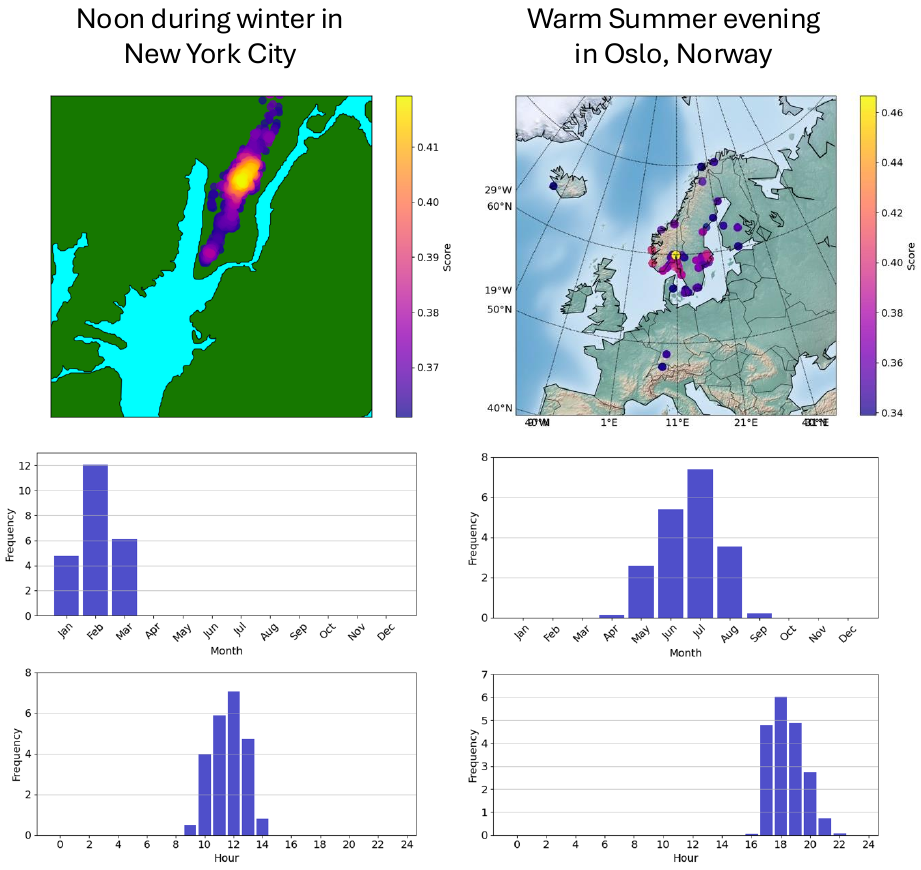}
  \caption{Qualitative examples of geo-localization and time-of-capture prediction using text queries. Top: prompt passed to CLIP's text encoder. Middle: spatial distribution of the predicted geo-locations with the highest cosine similarity. Bottom: histogram of the top-1k predicted months and hours with the highest cosine similarity. GT-Loc is capable of providing good estimates of the time and GPS coordinates to each of the text queries, even when time is not explicitly specified.}
  \label{fig:text}
\end{figure}

We also investigate the ability of GT-Loc to use the pretrained CLIP text encoder to retrieve times and locations mentioned in the text. For this task, we follow GeoCLIP's approach and replace the image backbone by a text backbone, keeping the trained MLP, location encoder and time encoder. For each text, we create a text embedding, pass it through the MLP and compare it against the location and time galleries. We then create spatial and temporal distributions of the top retrieved samples for each modality, as shown Figure \ref{fig:text}, where we see that not only is our model able to accurately pinpoint the location, but it also creates meaningful time distributions to words such as ``winter" and ``evening" that do not explicitly mention the time.
\vspace{-1em}

\vspace{-2em}
\section{Conclusion}
\label{sec:conclusion}

We introduce GT-Loc, a novel framework for jointly predicting the time and location of an image using a retrieval approach. GT-Loc not only shows competitive performance compared to state-of-the-art geo-localization models but also introduces the capability of precise time-of-capture predictions. A key innovation of our approach is the novel temporal metric loss, which significantly outperforms traditional contrastive losses in time prediction tasks. 

Furthermore, our results demonstrate that GT-Loc extends beyond standard time-of-capture prediction and geo-localization tasks. It supports additional functionalities like compositional image retrieval ($T+L \rightarrow I$), as well as text-to-location and text-to-image retrieval, indicating a profound understanding of the interplay between images, locations, and time.

\section{Acknowledgements}

Supported by Intelligence Advanced Research Projects Activity (IARPA) via Department of Interior/Interior Business Center (DOI/IBC) contract number 140D0423C0074. The U.S. Government is authorized to reproduce and distribute reprints for Governmental purposes notwithstanding any copyright annotation thereon. Disclaimer: The views and conclusions contained herein are those of the authors and should not be interpreted as necessarily representing the official policies or endorsements, either expressed or implied, of IARPA, DOI/IBC, or the U.S. Government.
{
    \small
    \bibliographystyle{ieeenat_fullname}
    \bibliography{main}
}

\clearpage
\maketitlesupplementary

% \section{Appendix}
% Optionally include supplemental material (complete proofs, additional experiments and plots) in appendix.
% All such materials \textbf{SHOULD be included in the main submission.}

% We organize the supplementary as follows. In Section~\ref{sec:m_infer} we present the inference pipeline of our framework, followed by training and implementation details in Section~\ref{sec:impl_supp} and dataset details in Section~\ref{sec:data_details}.

\section{Model inference}
\label{sec:m_infer}
\begin{figure}[h]
  \centering
  \includegraphics[width=\linewidth]{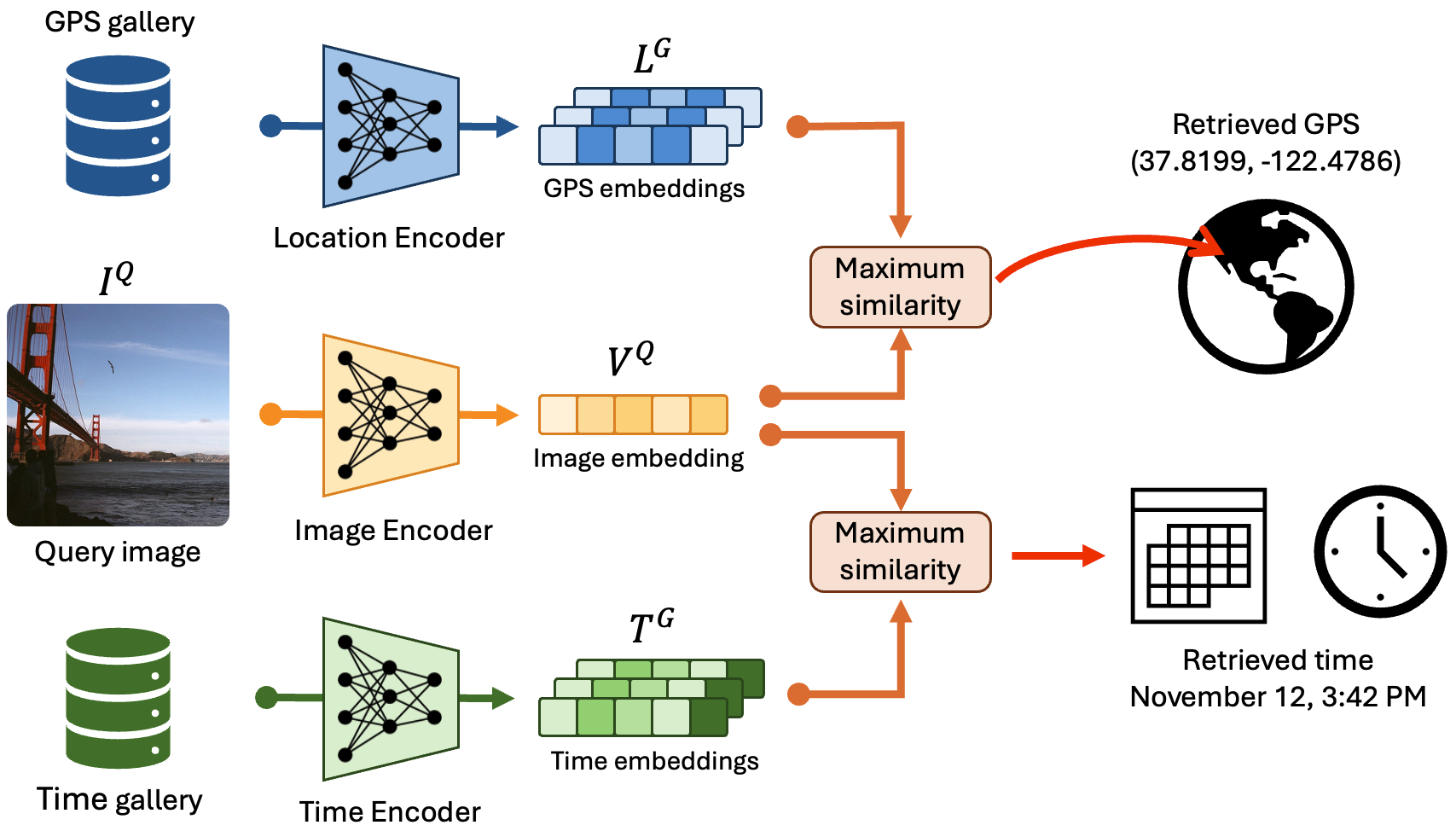}
  \caption{GT-Loc during inference.}
  \label{fig:inference}
\end{figure}

Our framework consists of a model that can predict both the location and capture-time of an image at the same time using a retrieval approach. Given a query image $I^Q$, a gallery of GPS coordinates and a gallery of timestamps, GT-Loc maps the three modalities into a shared feature space using an image, location and time encoder. The query image embedding $V^Q$ is compared against a set of location embeddings $L^G$ and time embeddings $T^G$. The GPS and timestamp with the highest cosine similarity are selected as the predictions of our model. Figure~\ref{fig:inference} represents the overview of our approach.

% \section{Limitations}
% Our current implementation of GT-Loc does not account for images from indoor scenes, which typically lack the distinct geo-temporal cues necessary for precise time prediction. 

% Although not included in this framework, a simple binary classifier could effectively distinguish between day and night images, serving as a preliminary filter to enhance our model’s applicability.

\section{Implementation details}
\label{sec:impl_supp}
Following GeoCLIP, the backbone of the image encoder is a pretrained ViT-L/14 from CLIP and the MLP consists of two fully connected layers with the ReLU activation function and dimensions 768 and 512 respectively. We use the same architecture for the time and location encoders as GeoCLIP. Both employ three RFF positional encoding layers, mapping the 2-dimensional GPS to a vector with 512 dimensions. The standard deviation values used to sample the RFF are $\sigma_i \in \{2^0, 2^4, 2^8\}$. The MLPs from the time and location encoder have three hidden layers with 1024 dimensions and a projection layer to map the final embeddings into a feature space of 512 dimensions. In the location encoder, we use a dynamic queue that stores the last 4096 seen locations, but we don't use it for time. The GPS coordinates and times are augmented by adding Gaussian noise with standard deviation of 150 meters for the in-batch GPS, 1500 meters for the GPS queue, 0.15 months and 0.15 hours for time. We perform two augmentations for each image in the training set using random resized crops of size 224, random horizontal flipping and image normalization. 

\section{Training protocol}
\label{sec:training_protocol}
GT-Loc is trained for 20 epochs using a cosine decay scheduler, with learning rate values ranging from $\alpha_{max}=3 \times 10^{-5}$ to $\alpha_{min}=3 \times 10^{-7}$. We use Adam optimizer with coefficients $\beta_1=0.9$, $\beta_2=0.999$ and $\ell_2$ penalty of $1 \times 10^{-6}$. For the contrastive losses, we use two learnable temperature parameters that are optimized during training. The batch size $B$ is set to 512 for all experiments, and the models are trained on a machine with 12 CPU cores and a NVIDIA RTX A6000 GPU.

\section{Additional qualitative results}

\begin{figure*}[h]
  \centering
  \includegraphics[width=\linewidth]{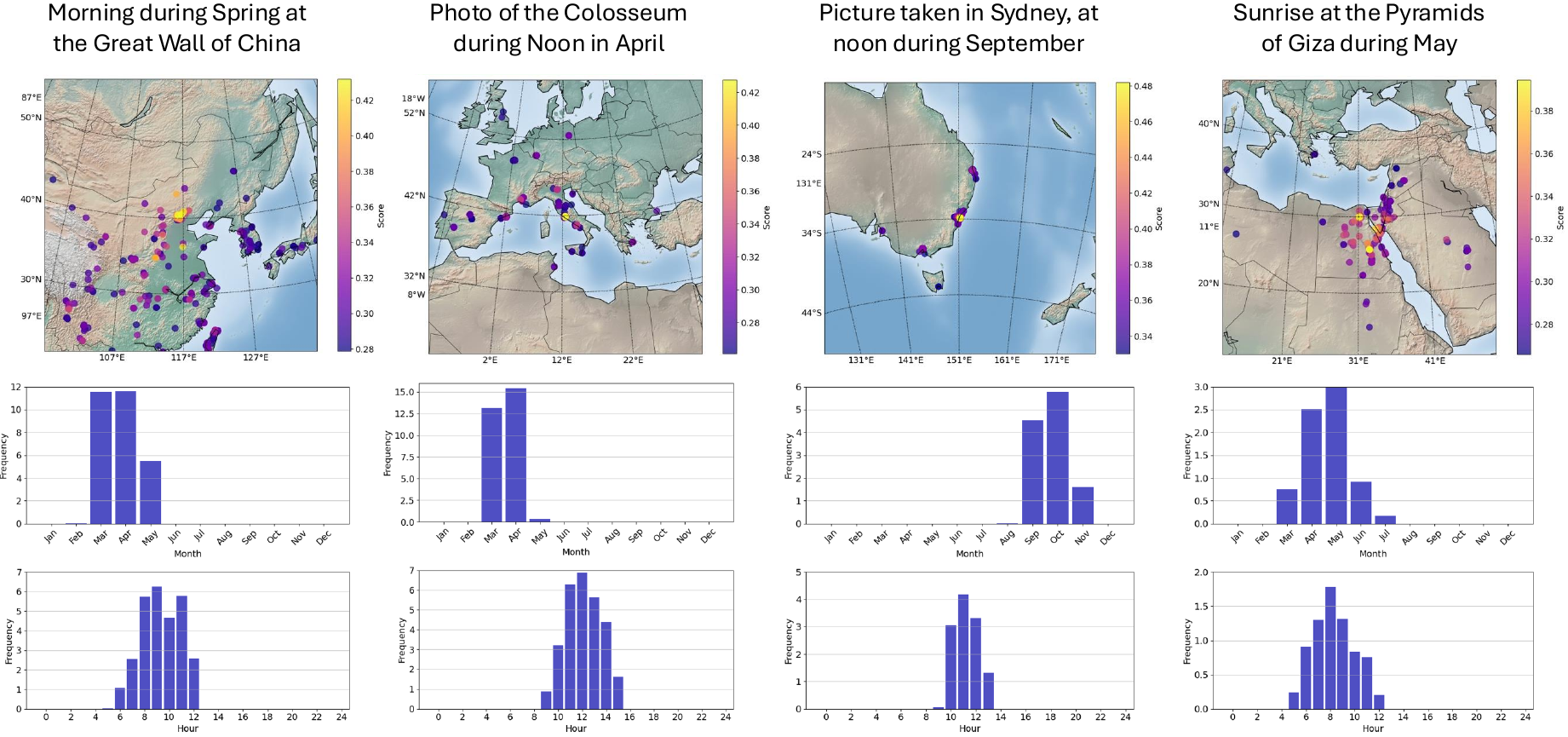}
  \caption{Additional qualitative examples of geo-localization and time-of-capture prediction using text queries. Top: prompt passed to CLIP's text encoder. Middle: spatial distribution of the predicted geo-locations with the highest cosine similarity. Bottom: histogram of the predicted months and hours with the highest cosine similarity.}
  \label{fig:supp-text}
\end{figure*}

\begin{figure*}
  \centering
  \includegraphics[width=0.75\linewidth]{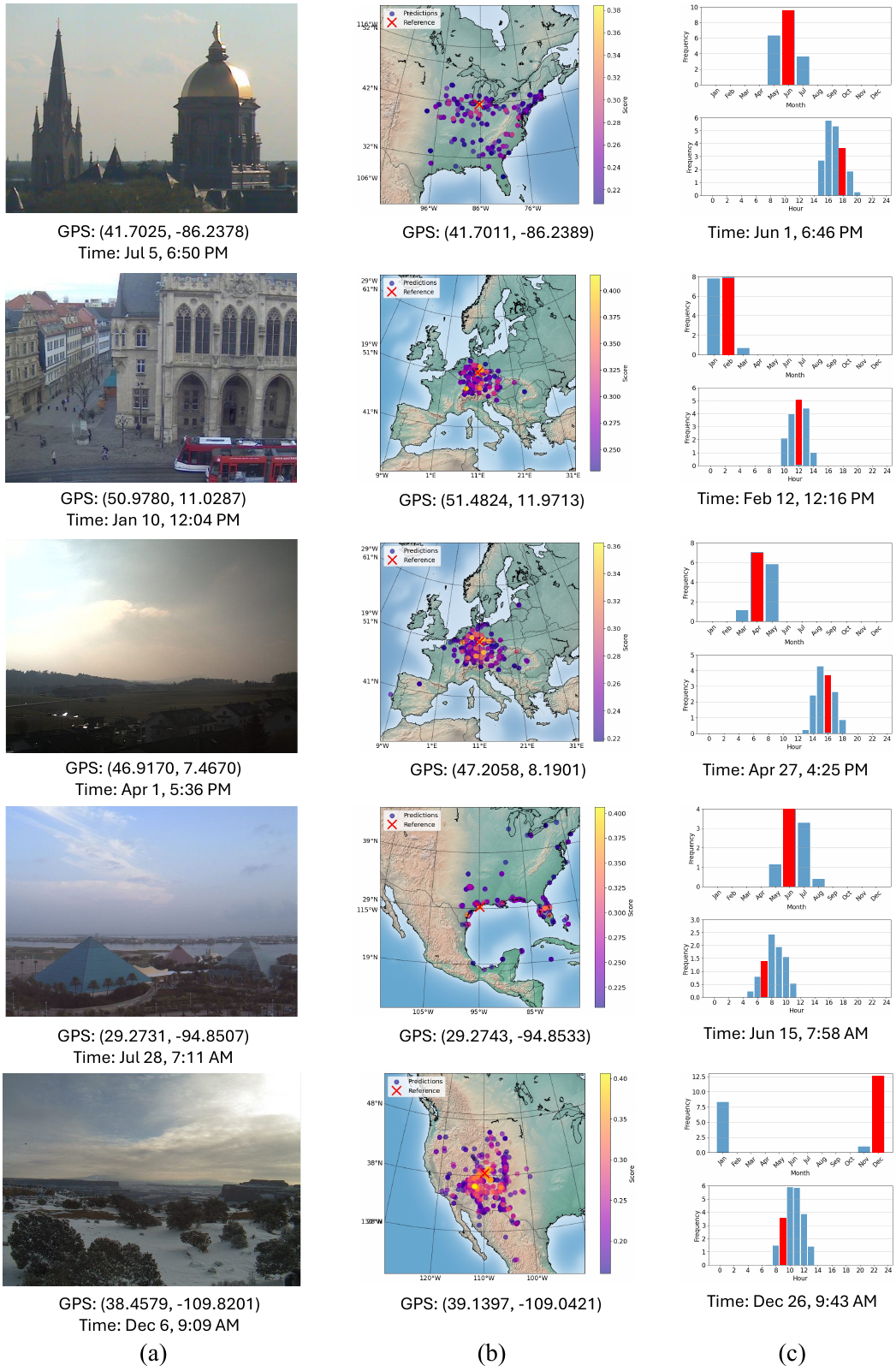}
  \caption{(a) Additional sample predictions for three cameras of the SkyFinder test set with the ground truth location and capture time. (b) Spatial distribution of the predicted GPS coordinates colored by the cosine similarity between the location and image embeddings. (c) Temporal distribution of the predicted month and hour, weighted by the cosine similarity between the time and image embeddings. The red bin contains the top-1 predicted time.} 
  \label{fig:more-qual}
\end{figure*}

We show additional qualitative results of our method in figures \ref{fig:supp-text} and \ref{fig:more-qual}. We include a failure case, on the last row of figure \ref{fig:more-qual}, where the time error is high because of the presence of fog in the image.

\section{Time-of-capture prediction histograms}

The time prediction histograms, shown in figures \ref{fig:qual}, \ref{fig:text}, \ref{fig:supp-text}, and \ref{fig:more-qual}, are computed using the following equation:

\begin{equation}
    \label{eq:supp_time_hist}
C_i = \sum_{j=1}^{N_G} \mathbbm{1}_{[j \in \mathcal{B}_i]} \cdot I^Q \cdot T^G_j,
\end{equation}

\noindent where $\mathcal{B}_i$ is the set of gallery embeddings that correspond to the $i$th bin, $C_i$ is the bin count, $N_G$ is the gallery size, $I^Q$ is the query image embedding, $T^G_j$ is the $j$th time embedding from the gallery, and $\mathbbm{1}$ is an indicator variable.

\section{Dataset details}
\label{sec:data_details}
We apply two filters to remove samples from the CVT that don't provide meaningful temporal information. In particular, we remove all night-time and indoor images, since they often have inconsistent temporal cues. To remove night images, we estimate the sunrise and sunset times from the date, latitude and longitude using the General Solar Position algorithm, and remove all samples before sunrise or after sunset. Then, for indoor images, we leverage a CNN model pretrained on the Places365 Dataset \cite{zhou2017places}. In general, night images often have inconsistent artificial lighting, more noise or specialized cameras such as night vision. Indoor images also have artificial lighting and controlled temperature, making it difficult to estimate the time or date. 

Regarding the levels of noise in the dataset, the SkyFinder subset consists of images with accurate time estimates, since they were collected from calibrated outdoor webcams. However, we observed that CVT has a moderate amount of noisy labels. Thus in order to train a model that can accurately predict the time, we need both datasets in the training set.

For evaluating the models, we employ a subset of unseen SkyFinder cameras, as well as two geo-localization datasets used by other state-of-the-art methods for evaluation: Im2GPS3k and GWS15k. Similar to GeoCLIP, we create a 100k GPS gallery to evaluate the model on Im2GPS3k, a 500k GPS gallery for GWS15k, and a 100k time gallery for the SkyFinder test set. The GPS galleries are created by sampling GPS coordinates from the MP-16 dataset, while the time gallery is created by sampling times from the combined CVT and SkyFinder training sets.

\section{Additional ablations}
\label{sec:ablations_supp}

\subsection{Image backbones}

To evaluate the impact of different image embeddings on time prediction performance, we conducted ablation studies using three backbones: DINOv2-L \citep{oquab2023dinov2}, OpenCLIP ViT-G \citep{ilharco_gabriel_2021_5143773}, and OpenAI's original CLIP ViT-L \citep{clip}. For these experiments, we used the TimeLoc variant of our model, which incorporates only the image and time encoders. The results, summarized in Table \ref{tab:img-encoders}, indicate that OpenAI's CLIP ViT-L achieves the lowest errors for both hours and months, as well as the highest Time Prediction Score (TPS).

\begin{table}[h!]
\centering
\begin{tabular}{lcccc}
\hline
\textbf{Backbone} & \textbf{Param.}       & \textbf{Month} & \textbf{Hour} & \textbf{TPS}  \\ 
{} & {} & \textbf{Error} & \textbf{Error} & {} \\
\hline
DINOv2-L & 0.3B                & 2.10                 & 3.25                & 68.71         \\ 
OpenCLIP-G & 1.8B              & 1.57                 & 2.94                & 74.65         \\
\textbf{CLIP-L} & \textbf{0.2B}           & \textbf{1.52}                 & \textbf{2.84}                & \textbf{75.49}         \\ \hline
\end{tabular}
\caption{Comparison of time prediction performance using different frozen image backbones.}
\label{tab:img-encoders}
\end{table}

\subsection{Time representation}
Motivated by the cyclical nature of time, \citealt{Aodha_2019_ICCV} used \textbf{circular decomposition} to wrap the temporal input to their geographical prior encoder, resulting in similar embeddings for dates that are close to the start and end of the year, such as December $31^{st}$ and January $1^{st}$. To achieve this,
for each dimension $l$ of the temporal input $\vec{x}$, they perform the mapping $[\sin(\pi x^l), \cos(\pi x^l)]$, resulting in two numbers for each dimension.

\textbf{Time2Vec} \citep{kazemi2019time2vec} is a method for encoding time that captures both periodic and non-periodic patterns. It transforms scalar time values into a vector of size \( k+1 \). The first element models linear, non-periodic trends, while the remaining elements are defined by a periodic activation function (e.g., sine), capturing repeating temporal behaviors like daily or weekly cycles. The representation is defined as:
\[
t2v(\tau)[i] =
\begin{cases} 
    \omega_i \tau + \phi_i & \text{if } i = 0, \\
    F(\omega_i \tau + \phi_i) & \text{if } 1 \leq i \leq k,
\end{cases}
\]
where \( F \) is a periodic activation function, typically \( \sin \), and \( \omega_i \) and \( \phi_i \) are learnable parameters representing the frequency and phase shift, respectively.

\begin{table}[ht]
\centering
\setlength{\aboverulesep}{0pt}
\setlength{\belowrulesep}{0pt}
\caption{\textbf{Ablations} for time prediction performance using different time encoders.}
\label{tab:time-encoder}
\begin{tabular}{lccc} 
\toprule
\textbf{Time}  & \textbf{Month} & \textbf{Hour} & \textbf{TPS} \\
\textbf{encoder} & \textbf{Error} & \textbf{Error} & \textbf{} \\
\midrule
Circular decomp. & 1.59 & 2.86 & 74.80 \\
Time2Vec          & 1.56 & 2.62 & 75.99 \\
RFF               & 1.40 & 2.72 & 77.00 \\
\bottomrule
\end{tabular}
\end{table}

\subsection{Time-of-year scale}
Time-of-year (ToY) can be represented at either a monthly or daily scale. In practice, the choice of time scale should not significantly affect the results, as the value is normalized before being passed to the time encoder, $\mathcal{T}(\cdot)$. However, two approaches are available. The first approach converts the integer month $m_i$ and day $d_i$ into a real-valued month, normalized over a 12-month period, as shown in Equations \ref{eq:time_rep_theta} and \ref{eq:time_rep_phi}. The second approach represents ToY as the number of days elapsed since the start of the year, normalized over 365 days (assuming no leap years in our dataset). This representation is defined as:
\[
\theta_i = \frac{1}{365} \left( d_i - 1 + \sum_{k=1}^{i} \mathcal{D}(m_k - 1) \right),
\]
where $\mathcal{D}(m_k)$ is the number of days in month $m_k$, and it is assumed that $\mathcal{D}(0)=0$. In Table \ref{tab:month2day}, we empirically show that using a monthly scale for ToY representation results in slightly better performance. However, we attribute this improvement to statistical noise rather than the time representation method itself.

\begin{table}[ht]
\centering
\setlength{\aboverulesep}{0pt}
\setlength{\belowrulesep}{0pt}
\caption{\textbf{Ablations} for time prediction performance using monthly and daily scales.}
\label{tab:month2day}
\begin{tabular}{lccc} 
\toprule
\textbf{ToY}  & \textbf{Month} & \textbf{Hour} & \textbf{TPS} \\
\textbf{Scale} & \textbf{Error} & \textbf{Error} & \textbf{} \\
\midrule
daily   & 1.45 & 2.71 & 76.61 \\
monthly & 1.40 & 2.72 & 77.00 \\
\bottomrule
\end{tabular}
\end{table}

\section{Compositional image retrieval details}
\label{sec:compositional}

To compare GT-Loc against a suitable baseline, we repurpose the model proposed by \citet{zhai2019learning} for compositional retrieval. In its original form, the model comprises an image encoder $C_I(I)$, a time encoder $C_T(t)$, and a location encoder $C_L(L)$. It concatenates the image and time embeddings to predict location via $P(l\mid C_I, C_T)$, and similarly concatenates the image and location embeddings to predict time via $P(t\mid C_I, C_L)$.

To adapt it for compositional image retrieval, we take a query time $T^Q$ and location $L^Q$ along with an image gallery ${I^G}$. First, we concatenate each image embedding $C_I(I^G)$ with $C_T(T^Q)$ or $C_L(L^Q)$ and feed these pairs to the respective classification heads. This yields probability distributions $P_i(l\mid C_I, C_T)$ and $P_i(t\mid C_I, C_L)$ for each image $i$ in the gallery, indicating how well that image matches the queried location and time. Since we already know the desired location and time, we extract the corresponding probabilities from the distributions and average them to produce a final similarity score. Ranking by this score allows us to retrieve the top-$k$ gallery images most likely to match $(T^Q, L^Q)$. This adaptation of the \citet{zhai2019learning} model provides a direct, fair comparison to GT-Loc's performance on compositional retrieval tasks.

\begin{figure}[h]
  \centering
  \includegraphics[width=0.8\linewidth]{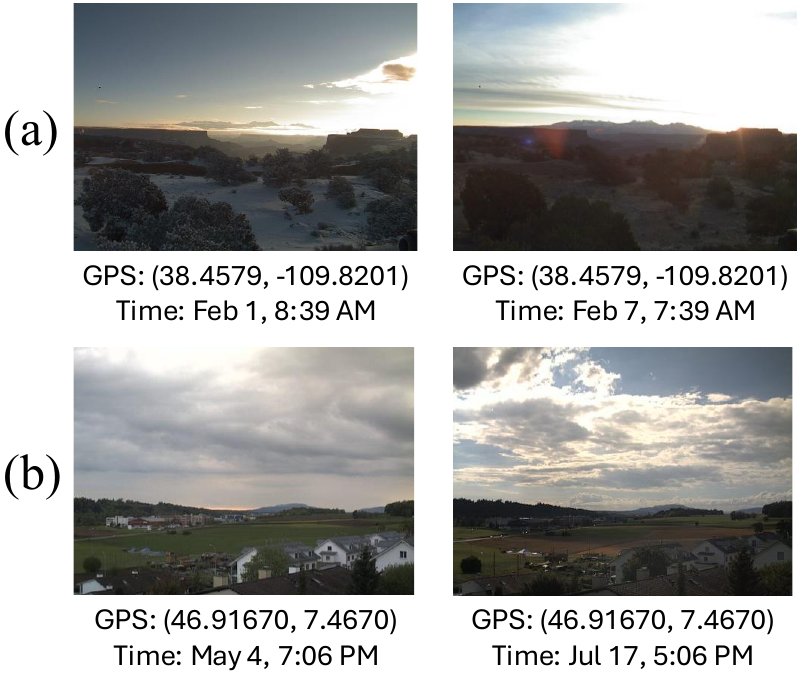}
  \caption{Illustrating \textit{compositional} $L+T\rightarrow{I}$ retrieval with GT-Loc. Each example Left: query location and query time, showing actual image. Right: retrieved image for given query location and time.
  } 
  \label{fig:comp-ret}
\end{figure}

Figure \ref{fig:comp-ret} shows two qualitative results of our compositional retrieval model.

\section{Additional geo-localization analysis}

\begin{figure}
  \centering
  \includegraphics[width=\linewidth]{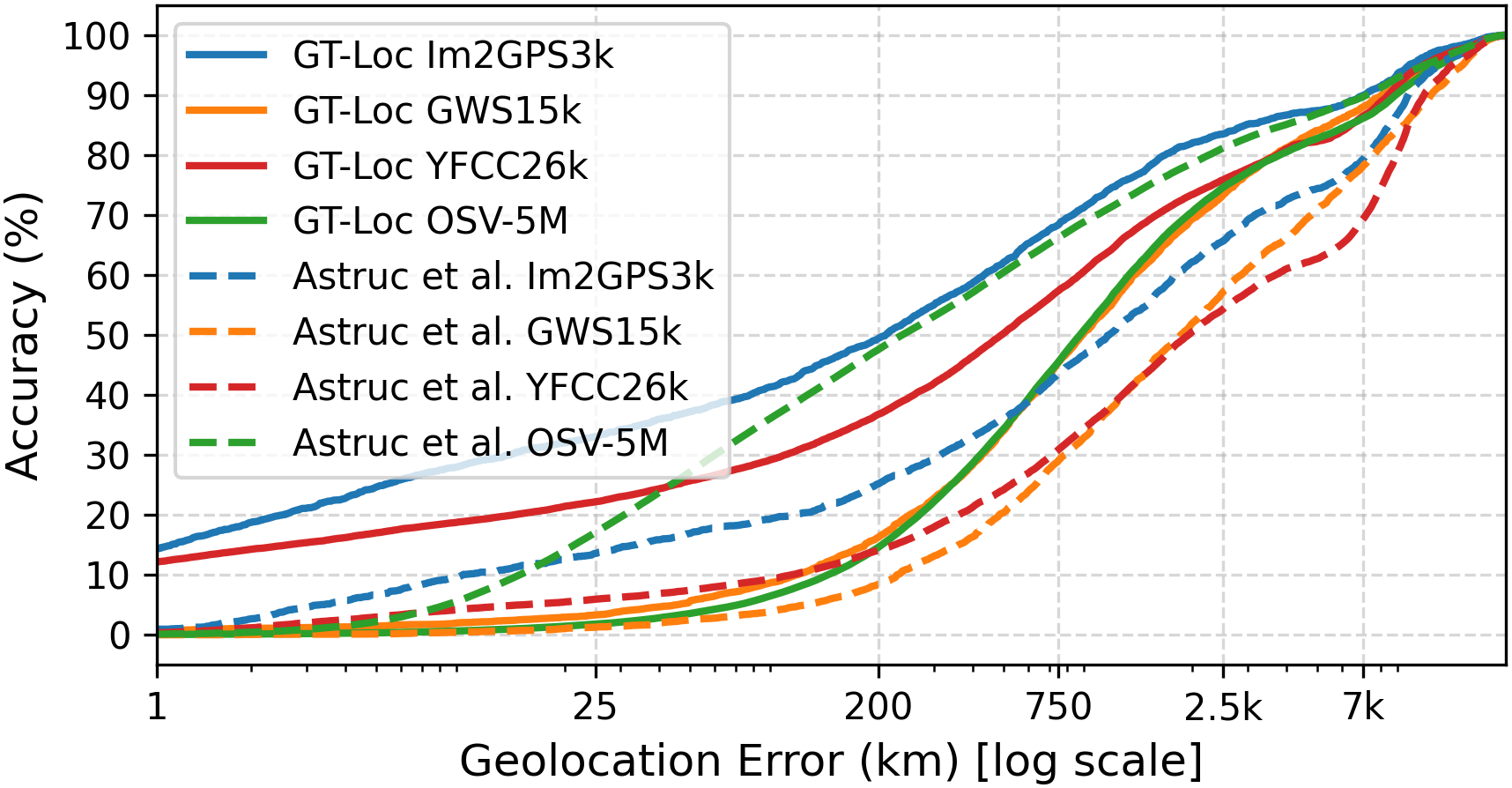}
  \caption{Cumulative geolocation error at different thresholds.}
  \label{fig:geoloc_thresholds}
\end{figure}

Figure~\ref{fig:geoloc_thresholds} presents the cumulative geolocation error evaluated over a range of distance thresholds. In addition to our main results, we include evaluations on two extra datasets: YFCC26k and OSV-5M. These plots allow for a more comprehensive comparison of model performance across diverse data distributions. We also include a comparison against the hybrid method proposed by \citet{osv5m}, which was not shown in the main paper. We observe that our method, GT-Loc, consistently outperforms \citet{osv5m} on all datasets with the exception of OSV-5M. We attribute this to the fact that OSV-5M contains imagery that is in-domain for their model, whereas it is out-of-domain for ours.

\section{Effect of population density on geo-localization}

Geo-localization datasets, particularly those collected via web-scraping or social media platforms, are inherently biased toward regions with higher population density. This raises a natural question: how does the performance of geo-localization algorithms vary between densely and sparsely populated areas? To investigate this, we conduct an additional analysis. For each image in our dataset, we obtain the corresponding population density from the Gridded Population of the World version 4 (GPWv4) dataset \citep{warszawski2017center}. We then group the data into ranges based on population density (measured in $pop/m^2$) and compute the median geo-localization error within each group. This stratified evaluation allows us to assess model performance across regions with varying levels of human activity. Consistent with the findings of \citet{osv5m}, we observe that geo-localization errors tend to be higher in areas with lower population density, as summarized in Table~\ref{tab:popdens-performance}.

\begin{table}[ht]
\centering
\caption{Median geo-localization error by population density group.}
\label{tab:popdens-performance}
\begin{tabular}{lcc} 
\toprule
      Pop. density & Im2GPS3k (km) & GWS15k (km)   \\
      \midrule
      $<100$                 & 266.66        & 1244.22       \\
      $[100,1\text{k})$      & 246.12        & 818.65        \\
      $[1\text{k},10\text{k})$ & 197.99      & 639.73        \\
      $\geq10\text{k}$       & 38.32         & 1035.81       \\
      \bottomrule
\end{tabular}
\end{table}

\section{Model performance vs. gallery size}

\begin{figure}[ht!]
  \centering
  \includegraphics[width=0.75\linewidth]{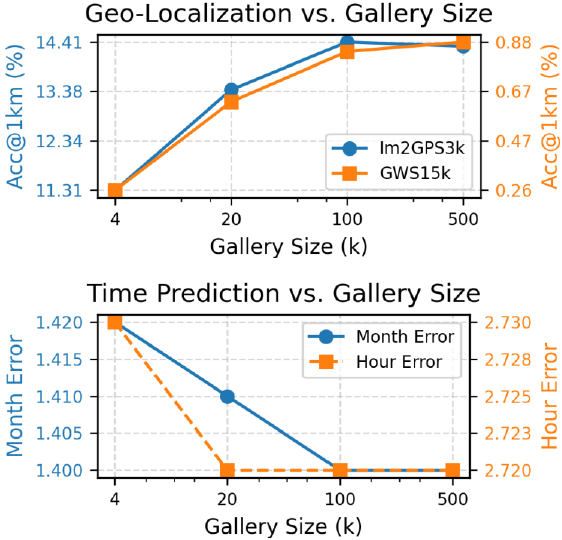}
  \caption{Effect of gallery size on geo-localization and time prediction.}
  \label{fig:perf_vs_gal}
\end{figure}

Figure~\ref{fig:perf_vs_gal} shows how the size of the gallery affects geo-localization and time prediction performance. We observe that increasing the size of the gallery leads to improved results, but the gains tend to saturate quickly. In particular, a gallery containing 100,000 samples is already sufficient to capture the majority of the performance improvements for geo-localization. Interestingly, for time-of-capture prediction, we find that even smaller archives, such as those with only 4,000 samples, can still yield strong performance.

\section{Analysis of the learned embedding space}

One of the key motivations for GT-Loc is to align images, time, and location in a shared multimodal embedding space. This approach is inspired by prior works like GeoCLIP \citep{vivanco2024geoclip}, SatCLIP \citep{klemmer2023satclip}, and CSP \citep{pmlr-v202-mai23a}, which embed images and GPS coordinates in shared spaces, as well as methods like ImageBind \cite{girdhar2023imagebind}, LanguageBind \citep{zhu2023languagebind}, Everything At Once \citep{shvetsova2022everything}, and Preserving Modality \citep{swetha2023preserving}, which align multiple modalities such as images, text, videos, and audio. Our work extends this idea to include temporal information, showing in Table \ref{tab:time-pred} that aligning these three modalities leads to improved time prediction performance compared to using only images and time.

\begin{figure}[h!]
  \centering
  \includegraphics[width=\linewidth]{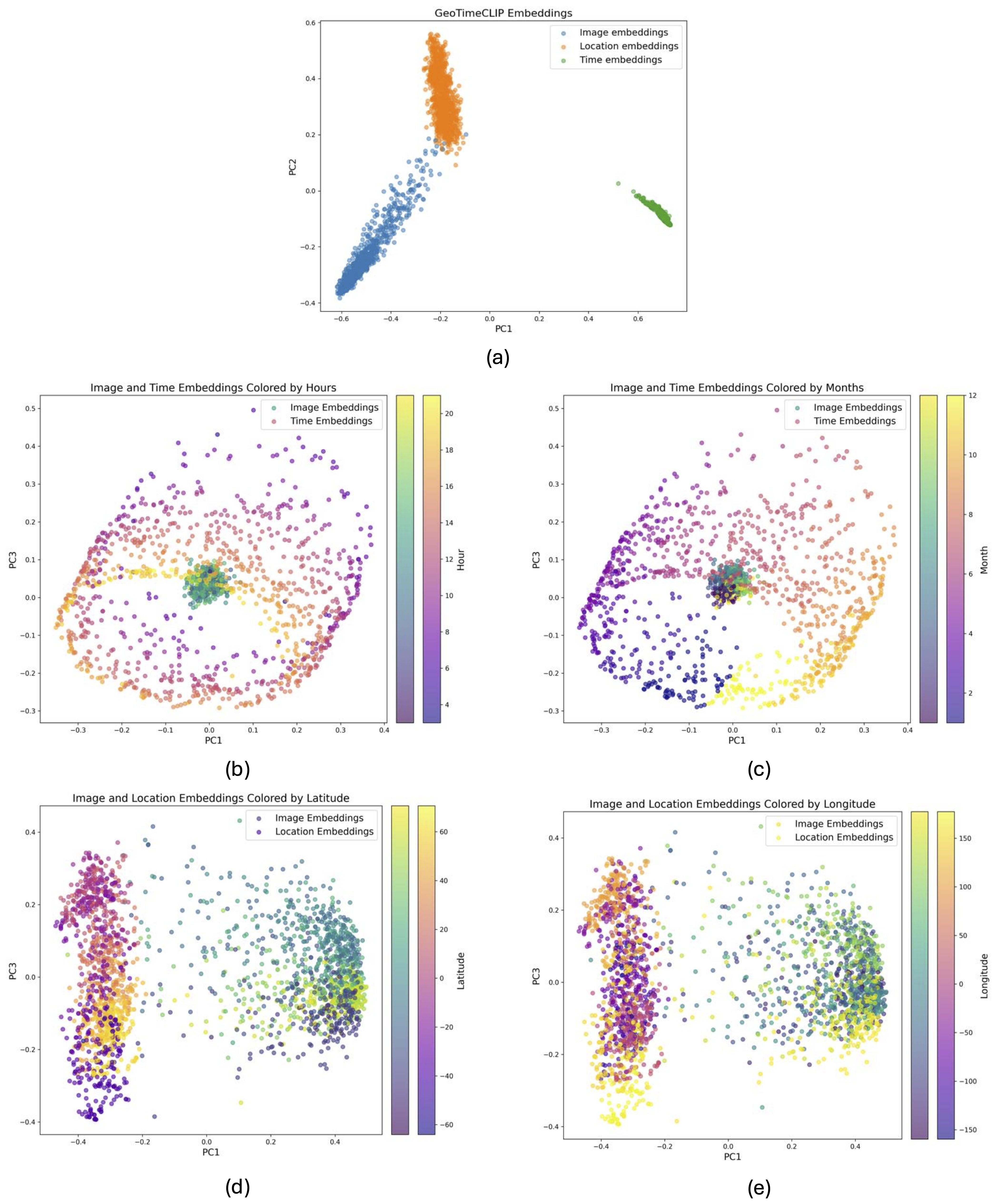}
  \caption{PCA plots of the embedding spaces in GT-Loc. (a) Distribution of the image, time and location embeddings. (b)-(c) Distribution of the image and time embeddings, colored by the time-of-day and time-of-year respectively. (c)-(d) Distribution of the image and location embeddings, colored by the latitude and longitude respectively.}
  \label{fig:pca}
\end{figure}

To explore the relationships between these modalities in the learned embedding space, we performed Principal Component Analysis (PCA) on the embeddings. While PCA has limitations in fully capturing the underlying structure of high-dimensional spaces, the results provide interesting qualitative insights. Figure \ref{fig:pca}(a) presents the distributions of image, time, and location embeddings, appearing in different subspaces. Figures \ref{fig:pca}(b) and \ref{fig:pca}(c) show more details about the relationship between image and time embeddings. The image embeddings are clustered in the center, surrounded by time embeddings. Notably, the directions of hours and months are well-defined: months are radially distributed, while hours are linearly distributed in a perpendicular direction. For location embeddings (Figures \ref{fig:pca}(d-e)), even though the patterns are less pronounced, the embeddings at different latitudes and longitudes still form distinct clusters in the feature space.

\section{Embedding distribution of non-overlapping panorama crops}

\begin{figure}[h]
  \centering
  \includegraphics[width=0.7\linewidth]{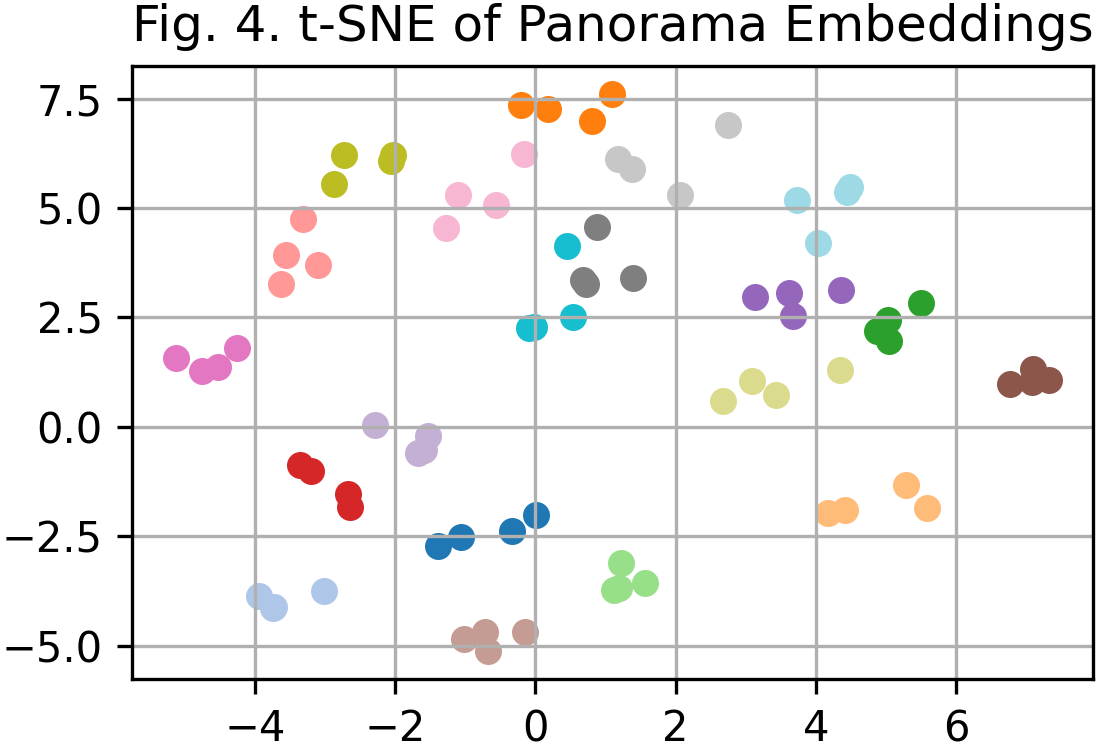}
  \caption{t-SNE visualization of image embeddings from non-overlapping $90^\circ$ crops of 20 panoramas sampled from the CVUSA dataset. Each color represents the four crops from a single panorama, which share the same time and location.}
  \label{fig:pano_embeddings}
\end{figure}

To further analyze the structure of our learned embedding space, we conduct a qualitative visualization using t-SNE (Figure \ref{fig:pano_embeddings}). We begin by randomly selecting 20 panoramas from the CVUSA dataset \citep{workman2015localize}. From each panorama, we extract four non-overlapping $90^\circ$ crops, resulting in a total of 80 image embeddings. Since all four crops from a single panorama share the same capture time and location by definition, they serve as a natural test of spatial and temporal consistency in the embedding space. In the resulting t-SNE plot, we observe that embeddings from the same panorama (shown using the same color) form tight and coherent clusters, suggesting that our model effectively encodes shared contextual information across views from the same scene.

\section{Scalability of the retrieval approach}

In Table~\ref{tab:params_mem_flops}, we present a comparison of the memory usage and computational cost (measured in FLOPs) between classification and regression baselines and our retrieval-based approach, across gallery sizes ranging from 4,000 to 500,000 samples. At the upper end of this range (500k), we observe that memory usage increases by approximately 950 MB. However, the additional computational cost introduced by the retrieval operations remains minimal, with only a +3.32\% increase in FLOPs. It is important to note that the gallery embeddings used for retrieval are precomputed offline in a single forward pass (requiring 9.47 TFLOPs) and are therefore not included in the runtime FLOPs reported in the table. 
% Beyond raw performance metrics, retrieval offers significant functional advantages: it enables capabilities that are inherently out of reach for classification or regression models, such as text-based retrieval and compositional retrieval involving multiple modalities.

\begin{table}[ht]
\centering
\caption{Memory usage and compute cost.}
\label{tab:params_mem_flops}
\setlength{\tabcolsep}{4pt}
\resizebox{\linewidth}{!}{
\begin{tabular}{l c c c c c c}
\toprule
Method & \multicolumn{3}{c}{Memory (GB)} & \multicolumn{3}{c}{TFLOPs} \\
\cmidrule(lr){2-4} \cmidrule(lr){5-7}
(gallery size)& Params & Gallery & Total & Forward & Retrieval & Total \\
\midrule
CLIP + cls    & 1.63 & --   & 1.63 & 159.41 & --     & 159.41 \\
CLIP + reg    & 1.63 & --   & 1.63 & 159.41 & --     & 159.41 \\
\midrule
GT-Loc (4k)   & 1.67 & 0.01 & 1.68 & 159.41 & 0.004  & 159.41 \\
GT-Loc (20k)  & 1.67 & 0.04 & 1.71 & 159.41 & 0.021  & 159.43 \\
GT-Loc (100k) & 1.67 & 0.19 & 1.86 & 159.41 & 0.105  & 159.52 \\
GT-Loc (500k) & 1.67 & 0.95 & 2.62 & 159.41 & 0.524  & 159.94 \\
\bottomrule
\end{tabular}
}
\end{table}

\section{Limitations and reproducibility challenges of existing time prediction methods}
\label{sec:supp_limitations}
Most previous time prediction methods suffer from a lack of standardization in their training and evaluation protocols. For instance, \citet{zhai2019learning} use subsets of the AMOS \citep{jacobs2007geolocating, jacobs09webcamgis} and YFCC100M \citep{thomee2016yfcc100m} datasets, without providing the source code or exact dataset splits necessary to replicate their experiments. Additionally, their time prediction evaluation relies on cumulative error plots, but they do not provide a single numerical value summarizing the performance of their model. Similarly, while \citet{salem2020learning} and \citet{padilha2022content} offer datasets and code, they do not include the cross-camera split for zero-shot time prediction—a more challenging and informative evaluation protocol that we adopt. Moreover, their results are presented only qualitatively, though they can be adapted for obtaining quantitative results. \citet{salem2022timestamp} also omit critical details such as dataset splits for the SkyFinder dataset, do not clarify whether their results corresponds to same- or cross-camera evaluation, and fail to provide the source code for replication. Their use of top-$k$ accuracy as an evaluation metric further complicates direct comparisons. Notably, none of these methods, with the exception of \citet{salem2020learning} and \citet{padilha2022content}, compare their time prediction performances against each other, and even these comparisons are only qualitative. Other time prediction approaches, including those by \citet{tsai2016photo}, \citet{li2017you}, \citet{lalonde2012estimating}, and \citet{hold2017deep}, also face similar challenges, such as a lack of available source code, missing datasets, or datasets that are no longer hosted online. This lack of standardization prevents consistent benchmarking across different methods.

% \section{Training stability}

% Maintaining training stability requires batches with consistent temporal and geographical distributions, which can be challenging with smaller batch sizes. To address this, we use batches of 512 randomly sampled images that span diverse locations, hours, and months. As shown in Figure \ref{fig:training-losses}, our batching strategy ensures smooth convergence of both the geo-localization loss ($\mathcal{L}_{loc}$) and the time metric loss ($\mathcal{L}_{time}$).

% \begin{figure}
%   \centering
%   \includegraphics[width=0.95\linewidth]{figures/training_losses.png}
%   \caption{\textcolor{blue}{Convergence of the geo-localization loss ($\mathcal{L}_{loc}$) and the Time Metric Loss ($\mathcal{L}_{time}$) during training.}} 
%   \label{fig:training-losses}
% \end{figure}

\section{Qualitative time-of-day results near sunrise and sunset}

\begin{figure}[h]
  \centering
  \includegraphics[width=0.8\linewidth]{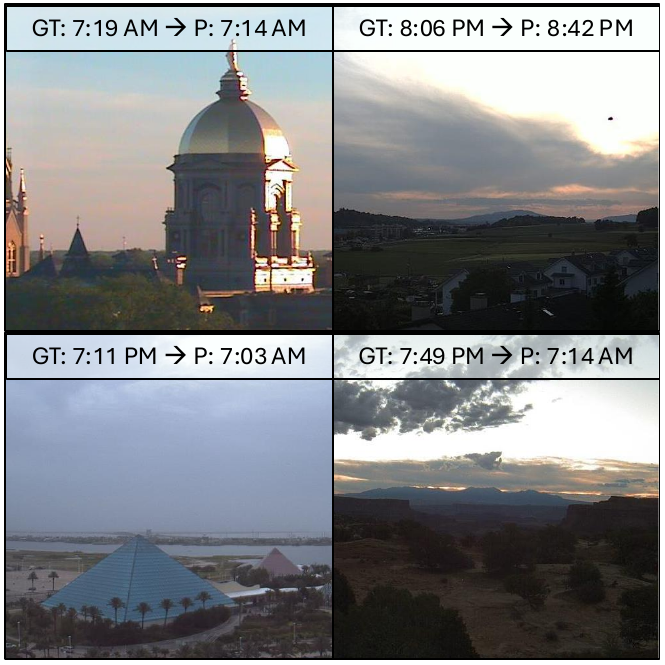}
  \caption{Qualitative time-of-day results near sunrise and sunset. Top row: correct predictions where the prediction (P) is within one hour of the ground truth (GT). Bottom row: failure cases.}
  \label{fig:sunrise-sunset}
\end{figure}

Predicting the time of day during periods close to sunrise and sunset is particularly challenging due to the visual similarity of scenes captured around these times. The task is further complicated by its strong dependence on geographic location (latitude) and the time of year (month), both of which have a direct influence on the time when these events occur. To illustrate these challenges, Figure~\ref{fig:sunrise-sunset} presents two qualitative examples where GT-Loc makes accurate predictions, as well as two examples where it fails.

\section{GT-Loc predictions across different latitudes}

\begin{figure}
  % \centering
  % \vspace{-15em}
  \includegraphics[width=\linewidth]{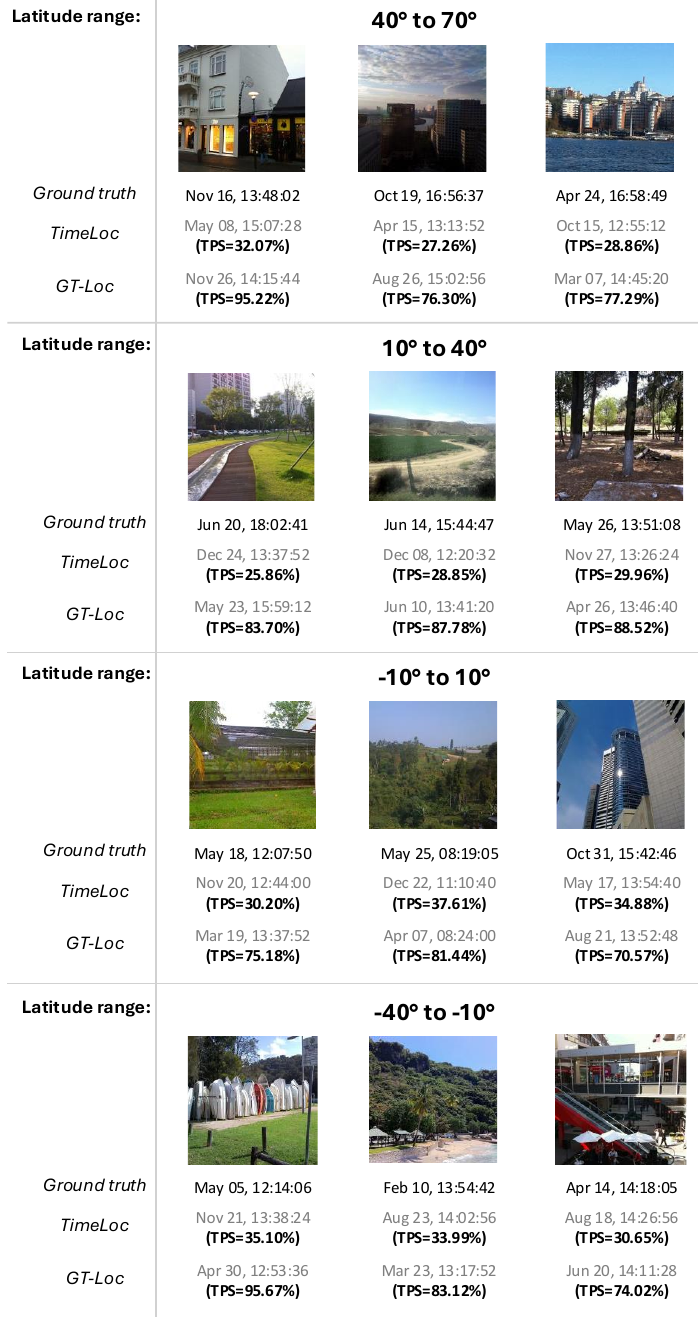}
  \caption{Sample predictions where GT-Loc outperforms the TimeLoc baseline across different latitudes.}
  \label{fig:timeclip-vs-geotimeclip}
\end{figure}

Figure \ref{fig:timeclip-vs-geotimeclip} compares time prediction examples from GT-Loc and TimeLoc, a baseline model trained solely with the visual ($\mathcal{V}$) and temporal ($\mathcal{T}$) encoders. The results suggest that TimeLoc struggles more with hour predictions at higher latitudes (40° to 70°) compared to GT-Loc. In contrast, at moderate latitudes (-40° to 40°), both models exhibit more consistent hour prediction errors, though GT-Loc demonstrates superior performance in month prediction.

\end{document}